\documentclass[10pt,twocolumn,letterpaper]{article}

\usepackage{cvpr}              % To produce the CAMERA-READY version
\definecolor{cvprblue}{rgb}{0.21,0.49,0.74}
\usepackage[pagebackref,breaklinks,colorlinks,allcolors=cvprblue]{hyperref}

\usepackage{orcidlink}
\usepackage{multirow}
\usepackage{booktabs}
\usepackage{threeparttable} 
\usepackage{color}
\usepackage{colortbl}

\usepackage{algorithm}
\usepackage{algpseudocode}

\usepackage{amsfonts}
\usepackage{bbm}
\usepackage{bm}
\usepackage{wrapfig,lipsum,booktabs}
\usepackage[utf8]{inputenc}
\usepackage{url}
\usepackage{booktabs}
\usepackage{amssymb}
\usepackage{bbding}
\usepackage{pifont}
\usepackage{wasysym}
\usepackage{utfsym}
\usepackage{amsmath}
\usepackage{amssymb}
\usepackage{bbm}
\usepackage{fontawesome}
\usepackage{wrapfig}

\usepackage{graphicx} 
\usepackage{pifont}
\usepackage[accsupp]{axessibility} 
\newcommand{\cmark}{\ding{51}} % 对勾
\newcommand{\xmark}{\textcolor{gray!40}{\ding{55}}} 

\definecolor{my_blue}{HTML}{d3eaf2}
\definecolor{Green}{rgb}{0.85882353, 0.90980392, 0.84705882}
\definecolor{rose}{rgb}{0.60392157, 0.53333333, 0.43921569}
\definecolor{dred}{rgb}{0.7254902, 0.09803922, 0.10588235}

\def\eg{\emph{e.g}\onedot} 
\def\ie{\emph{i.e}\onedot}

\newcommand{\upinc}[1]{\textcolor{red!60!black}{\scriptsize$\uparrow$\textbf{#1}}}

\def\paperID{666} % *** Enter the Paper ID here
\def\confName{CVPR}
\def\confYear{2026}

\title{FlowComposer: Composable Flows for Compositional Zero-Shot Learning
}

\author{Zhenqi He \qquad Lin Li \qquad Long Chen\textsuperscript{*} \\
 The Hong Kong University of Science and Technology\\
{\tt\small zheci@connect.ust.hk \qquad \{lllidy, longchen\}@ust.hk} \\
{\tt\small \url{https://hkust-longgroup.github.io/FlowComposer/}}
}

\begin{document}
\maketitle
\renewcommand{\thefootnote}{\fnsymbol{footnote}}
\footnotetext[1]{Corresponding author.}

\begin{abstract}
Compositional zero-shot learning (CZSL) aims to recognize unseen attribute–object compositions by recombining primitives learned from seen pairs. 
Recent CZSL methods built on vision-language models (VLMs) typically adopt parameter-efficient fine-tuning (PEFT).
They apply visual disentanglers for decomposition and manipulate token-level prompts or prefixes to encode compositions.
However, such PEFT-based designs suffer from two fundamental limitations: (1) Implicit Composition Construction, where composition is realized only via token concatenation or branch-wise prompt tuning rather than an explicit operation in the embedding space; (2) Remained Feature Entanglement, where imperfect disentanglement leaves attribute, object, and composition features mutually contaminated. Together, these issues limit the generalization ability of current CZSL models.
In this paper, we are the first to systematically study flow matching for CZSL and introduce FlowComposer, a model-agnostic framework that learns two primitive flows to transport visual features toward attribute and object text embeddings, and a learnable Composer that explicitly fuses their velocity fields into a composition flow. 
To exploit the inevitable residual entanglement, we further devise a leakage-guided augmentation scheme that reuses leaked features as auxiliary signals.
We thoroughly evaluate FlowComposer on three public CZSL benchmarks by integrating it as a plug-and-play component into various baselines, consistently achieving significant improvements. 
%
% Across three strong benchmarks for CZSL under both closed- and open-world protocols, \lil{FlowComposer consistently improves over strong CZSL baselines, demonstrating a scalable path to compositional generalization.}
\end{abstract}    
\section{Introduction}
\label{sec:intro}

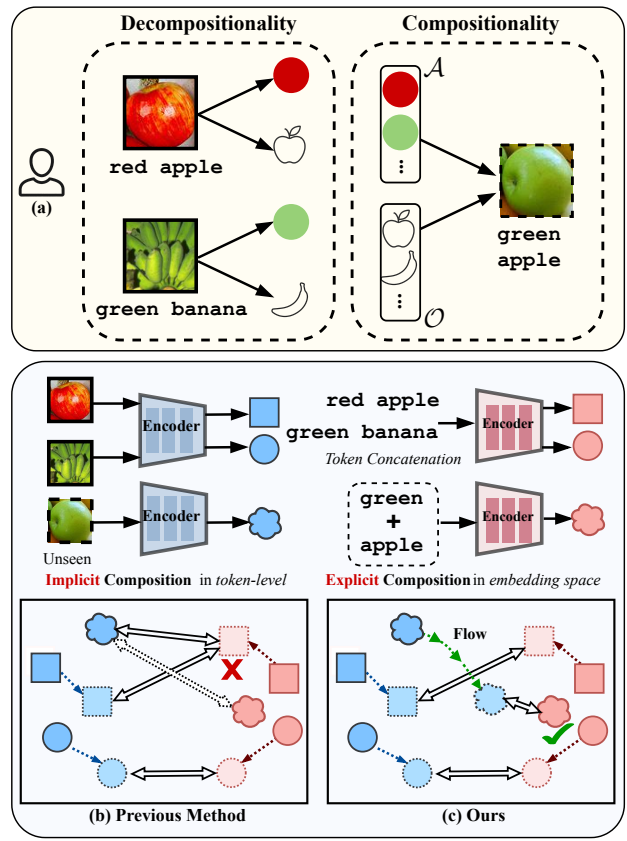
\begin{figure}[ht]
  \centering
  % 占位框：宽0.8\textwidth，高5cm
  \includegraphics[width=1\linewidth]{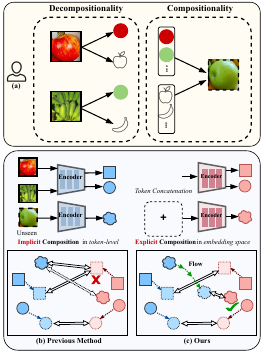}
  \put(-202,199.5){\scalebox{0.8}{\textbf{\texttt{green banana}}}}
  \put(-197,253){\scalebox{0.8}{\textbf{\texttt{red apple}}}}
  \put(-78,290){$\mathcal{A}$}
  \put(-78,195){$\mathcal{O}$}
  \put(-50,228){\scalebox{0.8}{\textbf{\texttt{green}}}}
  \put(-50,218){\scalebox{0.8}{\textbf{\texttt{apple}}}}
  \put(-115,165){\scalebox{0.8}{\textbf{\texttt{red apple}}}}
  \put(-130,153){\scalebox{0.8}{\textbf{\texttt{green banana}}}}
  \put(-102,128){\scalebox{0.8}{\textbf{\texttt{green}}}}
  \put(-102,110){\scalebox{0.8}{\textbf{\texttt{apple}}}}
  \caption{
(a) Humans recognize new concepts by recombining familiar primitives.
(b) Prior CZSL methods compose only at the token level, which may not yield valid unseen compositions in the embedding space.
(c) We perform explicit composition in the embedding space via learned attribute and object flows.
}
  \label{fig:teaser}
  \vspace{-10pt}
\end{figure}

%%%%% Background of CZSL
Humans exploit \emph{compositionality} and \emph{decompositionality}: we learn concepts by decomposing them into a set of primitives (\ie, attributes and objects) and recognize new concepts by recombining those primitives (\eg, from \textit{red apple} and \textit{green banana} to \textit{green apple} as shown in Fig.~\ref{fig:teaser}a). 
% from a small inventory of primitives, we routinely generate and recognize an unbounded set of novel concepts by recombining them (\eg., recognizing \textit{yellow apple} from \textit{yellow banana} and \textit{red apple}). 
%
Motivated by this, compositional zero-shot learning (CZSL)~\cite{nagarajan2018attributes,mancini2021open} is proposed to train on seen attribute-object compositions and test on unseen pairs formed from the same primitives. 
The core objective is therefore twofold: (i) Effectively learning decomposed primitive concepts from given images and (ii) Recomposing learned primitives into the composition representation for generalization.

% Humans exploit \emph{compositionality} and \emph{decompositionality}: we learn concepts by factoring them into a small set of primitives (attributes, objects) and recognize new concepts by recombining those primitives (\eg, from \textit{red apple} and \textit{yellow banana} to \textit{yellow apple}). 
% Motivated by this ability, \emph{compositional zero-shot learning (CZSL)} trains on \emph{seen} attribute--object pairs and requires recognizing \emph{unseen} pairs formed from the same primitive sets. 
% The core objective is therefore twofold: (i) learn primitives that can be \emph{decomposed} from data and (ii) provide an \emph{explicit operation} that \emph{composes} primitives into reliable composition representations for recognition.

%%%%% Recent methods & Challenges
Recent CZSL approaches build on powerful vision-language models (VLMs) like CLIP~\cite{radford2021learning}, capitalizing on their aligned visual-textual spaces pre-trained on large-scale image-text pairs. 
In general, they measure the similarities between the query image and the text embedding of each primitive concept for zero-shot recognition.
To better leverage the compositional nature, recent efforts combine both decomposed and composed strategies, powered by parameter-efficient fine-tuning (PEFT), \eg, text prompt tuning and visual adapter.
% To better leverage the compositional nature, recent efforts leverage parameter-efficient fine-tuning (PEFT), \eg, text prompt tuning and visual adapter, with both decomposed and composed strategies. 
%
\textit{On the decomposition side}, two visual disentanglers are employed to independently extract visual features for attributes and objects~\cite{huang2024troika,lu2023decomposed}. These features are then used to compute classification logits against their respective primitive text embeddings.
\textit{On the composition side}, compositional semantics are captured by constructing token-level text prompts to bind attributes and objects, and align the encoded text features with their corresponding image features.
For instance, CSP~\cite{nayaklearning} concatenates learnable \texttt{[attribute]} and \texttt{[object]} tokens within one prompt with fixed prefix (\texttt{a photo of}), while \cite{lu2023decomposed,huang2024troika} extend with branch-specific soft prefixes, implicitly modeling different semantic roles at the token level.
% the composition category semantic features, derived from the concatenated \texttt{[attribute]} \texttt{[object]} tokens, are directly aligned with the holistic visual features of the image~\cite{nayaklearning,huang2024troika}. 
% In summary, the parameter-efficient fine-tuning (PEFT) strategies, \eg, text prompt tuning and visual adapter, are widely used to efficiently adapt VLMs for CZSL task.

% To adapt VLMs to CZSL, methods typically align compositional visual and textual features \lil{through parameter-efficient fine-tuning, \eg, text prompt tuning and visual adapter~\cite{nayaklearning,purushwalkam2019task,lu2023decomposed}.}
% %
% Decompositionality is typically pursued by visual disentanglers that split compositional features into attribute/object streams, while compositionality is approximated by concatenating \texttt{[attribute]} \texttt{[object]} tokens in the text branch. 

\begin{figure}[t]
  \centering
  % 占位框：宽0.8\textwidth，高5cm
  \includegraphics[width=1\linewidth]{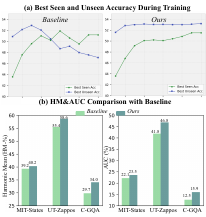}
  \vspace{-2em}
  \caption{
Training dynamics and performance comparison with baseline - Troika~\cite{huang2024troika}. Our method yields a more balanced seen/unseen accuracy trajectory and consistently improves HM and AUC over the baseline on all three datasets.
}
  \label{fig:comp}
  \vspace{-10pt}
\end{figure}

However, these PEFT-based decomposition-composition strategies suffer from two fundamental flaws: 1) \textbf{Implicit Composition Construction}: 
The composition relies on \emph{token-level} concatenation rather than an explicit operation in the embedding space, so for unseen pairs the resulting composition embedding may not lie close to the corresponding image embedding, as illustrated in Fig.~\ref{fig:teaser}b, which in turn hampers generalization to unseen compositions.
2) \textbf{Remained Feature Entanglement}: The visual disentangler fails to enforce strict feature separation for primitives, resulting in visual streams leaking information across primitives. Such residual entanglement weakens branch-specific alignment and makes recomposed embeddings for unseen attribute--object pairs less reliable.
Taken together, these flaws make PEFT-based CZSL methods prone to overfitting, and struggle to achieve a stable trade-off between seen and unseen recognition ability.
As shown in Fig.~\ref{fig:comp}a (left), as training progresses, the seen accuracy rises, whereas the unseen accuracy steadily drops, exposing a strong bias toward seen compositions.

Inspired by the recent success of Flow Matching (FM) where time-conditioned velocity models robustly transport latents across spaces for image generation~\cite{liu2022flow,lipman2023flow,albergo2023building,liu2025flowing,he2025flowtok},
%
% we posit that the same mechanism is a natural fit for CZSL.
we explore flow matching as a natural computational mechanism for CZSL.
The core insight is that FM's velocity fields naturally admit combination and decomposition, aligning naturally with the essence of compositionality and decompositionality in CZSL.
%
% The core insight is that FM's velocity fields are inherently composable and decomposable, aligning naturally with the essence of compositionality and decompositionality in CZSL.
% where velocity fields are inherently composable and decomposable, aligning naturally with the essence of compositionality and decompositionality. \lil{The core insight is that FM's velocity fields are inherently composable and decomposable, aligning naturally with the essence of compositionality and decompositionality in CZSL.}
%
Guided by this view, we propose \textit{FlowComposer}, a flow-based CZSL framework that first learns two primitive flows to transport visual features toward their corresponding text embeddings, and then explicitly composes these primitive velocity fields to form composition flows in the shared embedding space, as illustrated in Fig.~\ref{fig:teaser}c.

To address the \emph{implicit composition construction}, we introduce an explicit \textbf{\emph{Composer}} in the embedding space that combines primitive velocities into a composition velocity, yielding a learnable composition rule for transporting between compositional text and image embeddings from primitive embeddings.
% To address the \emph{implicit composition construction}, we introduce an explicit \textbf{\emph{Composer}} in the embedding space that combines the primitive velocities into a composition velocity, yielding a learnable composition rule for transporting between compositional text and image embeddings from primitive embeddings.
%
% providing an operational rule for transporting between compositional text and image embeddings based on primitive embeddings. 
%
% To tackle the \emph{remained feature entanglement}, we further design a \textbf{\emph{leakage-guided augmentation}}: beyond standard within-branch supervision (\eg, attr-image $\rightarrow$ attr-text), we repurpose leaked features as cross-branch signals (\eg, attr-image $\rightarrow$ obj-text), enriching velocity supervision and making the learned flows more robust to imperfect disentanglement.
%
To address the \emph{remained feature entanglement}, we introduce a \textbf{\emph{leakage-guided augmentation}}: instead of treating cross-branch leakage as a failure of disentanglement, we explicitly reuse leaked features as cross-branch supervision signals (\eg, attr-image $\rightarrow$ obj-text) alongside standard within-branch supervision, thereby enriching velocity supervision and turning imperfect disentanglement into a source of guidance.
Crucially, \textit{FlowComposer} is model-agnostic and plug-and-play: it operates purely in the representation space, aims to tighten the distance between composed text and image embeddings, and can therefore be attached to diverse CZSL pipelines with only minor modifications.
%
% We adapt our framework to popular CZSL baselines, CSP~\cite{lu2023decomposed} and Troika~\cite{huang2024troika}, obtaining substantial HM\&AUC gains and a more stable, balanced trade-off between seen and unseen compositions as shown in Fig.~\ref{fig:comp}b.
% We adapt our framework to popular CZSL baselines, CSP~\cite{lu2023decomposed} and Troika~\cite{huang2024troika}, obtaining substantial HM\&AUC gains and a more stable, balanced trade-off between seen and unseen compositions, as shown in Fig.~\ref{fig:comp}, where our curves are steadier and bars consistently higher.
% We adapt our framework to popular CZSL baselines, CSP~\cite{lu2023decomposed} and Troika~\cite{huang2024troika}, obtaining substantial HM\&AUC gains and a more stable, balanced trade-off between seen and unseen compositions, as shown in Fig.~\ref{fig:comp}b, where FlowComposer consistently improves HM and AUC across datasets.
We adapt our framework to popular CZSL baselines, CSP~\cite{nayaklearning} and Troika~\cite{huang2024troika}, leading to higher HM and AUC and a more stable balance between seen and unseen compositions, as shown in Fig.~\ref{fig:comp}, where \textit{FlowComposer} reduces the seen–unseen bias and improves overall performance across all datasets.

% we replace static proximity with explicit transport in the embedding space by learning primitive-specific flows whose dynamics can be composed into a composition flow. 
%
% To leverage flow matching to bridge between compositional text and image embeddings and encode the composition operation, 
% We propose a simple yet effective framework \textit{FlowComposer}, a two-path flow framework that separately learns attribute and object velocity fields, together with a learnable composer that combines them into a composition velocity for the transportation between compositional text and image embeddings.
% %
% To exploit the inevitable leakage in visual disentanglement, we further introduce a leakage-guided augmentation within \textit{FlowComposer}. Beyond the standard within-branch supervision (\eg, attr-image $\rightarrow$ attr-text), we re-purpose leaked cues as auxiliary cross-branch targets (\eg, attr-image $\rightarrow$ obj-text), thereby enriching the velocity supervision and strengthening composition transport.
% %
% Crucially, \textit{FlowComposer} is model-agnostic and plug-and-play: it operates purely in the representation space, aims to tighten the distance between composed text and image embeddings, and therefore can be overlaid on diverse CZSL pipelines with only minor changes.
% %
% We adapt our framework to popular CZSL baselines - CSP~\cite{lu2023decomposed} and Troika~\cite{huang2024troika}, obtaining substantial improvements for them.

In summary, we make the following contributions in this paper:
\textbf{(i)} We take an initial step towards exploiting flow matching for CZSL, and propose \textit{FlowComposer}, a general framework that employs composable flows atop existing CZSL pipelines to encode explicit composition operations.
\textbf{(ii)} We introduce a leakage-guided augmentation that leverages the inherent imperfect disentanglement of attribute and object features, converting cross-branch leakage into constructive supervision to boost composition recognition.
\textbf{(iii)} Extensive experiments on public CZSL datasets show that plugging \textit{FlowComposer} into diverse CZSL pipelines consistently improves performance.

\section{Related Work}
\label{sec:related_work}

\paragraph{Compositional Zero-Shot Learning (CZSL).}
CZSL seeks to recognize \emph{unseen} attribute–object compositions by leveraging supervision from \emph{seen} compositions and recombining primitive concepts learned from labels of the form (attribute, object). Early approaches fall into two broad lines. The first learns a single classifier or shared embedding that maps images and compositions into a common space for direct composition prediction, typically enforcing relations between primitives and compositions ~\cite{misra2017red,cgqa,nagarajan2018attributes,li2020symmetry,liu2023simple}. The second decouples attribute and object modeling with parallel heads, then fuses the two streams for composition recognition via late fusion ~\cite{purushwalkam2019task,hao2023learning,kim2023hierarchical,yang2020learning,wang2023learning,karthik2022kg}.

With the rise of powerful vision–language models (VLMs)~\cite{radford2021learning}, CZSL methods increasingly adopt prompt-based paradigms that exploit pre-trained alignment between images and text~~\cite{nayaklearning,lu2023decomposed,zhang2025learning,huang2024troika}. 
CSP~\cite{nayaklearning} uses a \emph{single-path} design that represents the whole composition as one learnable prompt token and aligns this single branch with image features. In contrast, later methods~\cite{lu2023decomposed,huang2024troika,zhang2025learning,li2025texth2emlearninghierarchicalhyperbolic} adopt a \emph{multi-path} formulation that splits the text stream into attribute and object branches with branch-specific prompts, aligns each branch to visual features, and then fuses the branches for prediction.
Recent work goes further by using large language models (LLMs) to \emph{generate additional textual knowledge} that improves compositional reasoning. For example, PLID~\cite{PLID} prompts LLMs to generate diverse, informative class descriptions to form language-informed text distributions, LOGICZSL~\cite{wu2025logiczsl} converts LLM-derived relational knowledge into differentiable logic rules as auxiliary losses, and PLO-LLM~\cite{li2023compositional} uses LLMs to craft composition-specific graduated descriptions that guide progressive, step-wise observation.

\paragraph{Flow Matching (FM).}
FM has recently emerged as a leading paradigm for image generation.
It learns a time-conditioned velocity field (neural ODE) that transports samples along a prescribed probability path from an easy prior to the data distribution by directly regressing the target velocity~\cite{lipman2023flow,albergo2023building,liu2022flow}.
Early work~\cite{esser2024scaling,geng2025mean} applies this transport from noise priors to images, yielding high-fidelity synthesis comparable to diffusion~\cite{ho2020denoising} but with simpler dynamics.
More recently, text-to-image FMs~\cite{liu2025flowing,he2025flowtok} learn cross-modal flows that carry text embeddings to image embeddings, enabling direct transportation between language and vision via learned velocities.
Beyond text-to-image generation, FM has been extended to image-level transfers, \eg, depth estimation~\cite{gui2025depthfm} and semantic segmentation~\cite{bogensperger2025flowsdf,wang2025deforming}.

\paragraph{Generative Models for Classification.} Beyond synthesis, modern generative models have been widely adopted across visual perception and classification~\cite{wang2025inversion,wang2025noise,liu2025generate,clark2023text,li2023your,chen2023robust,qi2024simple}. A large thread uses generative models to augment scarce data by synthesizing faithful and diverse samples that expand training sets for fine-grained classification~\cite{islam2024diffusemix, fu2024dreamda}, few-shot~\cite{wang2025inversion}, long-tailed~\cite{koh2025synthetic}, and category discovery~\cite{liu2025generate}.
A complementary line treats the generator itself as a classifier, converting text-to-image diffusion models into zero-shot and few-shot generative classifiers by scoring class-conditional denoising (a “diffusion classifier”)~\cite{clark2023text,li2023your,chen2023robust} and improving efficiency with hierarchical prompting~\cite{ning2024hierarchical} and additional image encoder~\cite{qi2024simple}. Recent work also parameterizes few-shot learners via diffusion time-steps to isolate nuanced class attributes~\cite{yue2024few}, and optimizes the matching noise to stabilize the pipeline~\cite{wang2025noise}.

While the above approaches all leverage pretrained text-to-image generators, they primarily use them either for data synthesis or by repurposing the generator as a classifier. In contrast, we exploit the core principle of distribution transportation and utilize the compositionality of velocity fields.
Notably, two concurrent works~\cite{jiang2025exploring,li2026path} apply flow matching for few-shot cross-modal alignment, but they do not exploit the compositional capacity of velocity fields.
\section{Preliminary}
\label{sec:pre}
In this section, we first elaborate on the CZSL task in Sec~\ref{sec:pre:ps}, offer a review of baselines in Sec~\ref{sec:pre:baselines}, and introduce the flow matching mechanism in Sec~\ref{sec:pre:fm}.

% Task Definition
\subsection{Problem Statement}
\label{sec:pre:ps}
CZSL aims to recognize unseen attribute–object compositions by composing primitive concepts learned from seen compositions.
Consider an attribute set $\mathcal{A}=\{a_1,\dots,a_M\}$ and an object set
$\mathcal{O}=\{o_1,\dots,o_N\}$, the universe of compositional classes is the
Cartesian product $\mathcal{C}=\mathcal{A}\times\mathcal{O}$, where each class
$c=(a,o)$. $\mathcal{C}$ is partitioned into disjoint seen and unseen subsets,
$\mathcal{C}_s$ and $\mathcal{C}_u$, with $\mathcal{C}_s\cap\mathcal{C}_u=\varnothing$.
The training set is $\mathcal{D}_{\mathrm{tr}}=\{(I_i,c_i)\mid I_i\in\mathcal{I},~c_i\in\mathcal{C}_s\}$, where $\mathcal{I}$ denotes the image domain.
At test time, the model predicts over a target composition space
$\mathcal{C}_{\mathrm{test}} \subseteq \mathcal{C}$ under two standard protocols: (i) \textbf{Closed-world:} $\mathcal{C}_{\mathrm{test}}=\mathcal{C}_s \cup \mathcal{C}_u^{\mathrm{sub}}$ for some
$\mathcal{C}_u^{\mathrm{sub}} \subseteq \mathcal{C}_u$ where $\mathcal{C}_u^{\mathrm{test}}$ is the unseen compositions in test data; (ii) \textbf{Open-world:} $\mathcal{C}_{\mathrm{test}}=\mathcal{C}$ where the candidate label set is all valid compositions.

% Baseline Method
% 1, One-path baseline - CSP
% 2, 3-path baseline - Troika
\subsection{Review of CZSL Baselines}
\label{sec:pre:baselines}
\noindent\textbf{Single-path Baseline.}
CSP~\cite{nayaklearning} formulates CZSL as a single-path alignment between the composition text embedding and the image embedding. Specifically, it freezes both CLIP encoders and optimizes only learnable text input tokens for attribute and object, which are inserted into a fixed template (``a photo of \texttt{[attribute]} \texttt{[object]}''). 
For each seen pair \((a,o)\), the predicted probability is computed by applying a softmax over temperature-scaled cosine similarities between the image feature and the prompt embedding, and the training objective is to minimize the cross-entropy with \(\,L_2\) regularization on the learnable text tokens.

% \[
% \mathcal{L} = -\frac{1}{|S_{\text{seen}}|}\!\sum_{(x,y)\in S_{\text{seen}}}\log p_{\theta}(y\mid x)\;+\;\lambda \lVert \theta \rVert_2^2,
% \]

\noindent\textbf{Multi-path Baseline.}
% \cite{huang2024troika} introduces a three-path framework to align branch-specific embeddings jointly. Both CLIP encoders are frozen. On the vision side, it employs a lightweight trainable visual adapter with a disentangler to separate branch-specific visual features. On the text side, it applies a shared primitive vocabulary with a learnable soft token for state and object, and branch-specific soft prefixes for the state/object/composition prompts. To further align closer the embeddings, it introduces a cross-modal traction module to use cross-attention to inject diverse visual content into the prompt representation.
% %
% Given an image $x$ and a pair $(s,o)$, each branch encodes its prompt and scores $x$ with temperature-scaled cosine similarity, yielding $p_s(s\mid x)$, $p_o(o\mid x)$, and $p_c((s,o)\mid x)$.
% %
% Its training objective is to minimize the sum of cross-entropy losses over seen state, object, and pair branches. During inference, it computes the predicted probability as a calibrated product $p_c\!\cdot\!p_s\!\cdot\!p_o$ over branches.
Troika~\cite{huang2024troika} proposes a three-path architecture that jointly aligns branch-specific embeddings for state, object, and composition. Both CLIP encoders are kept frozen. On the vision side, it introduces a lightweight trainable adapter together with respective disentanglers to extract branch-specific visual features. On the text side, it employs the same learnable primitive soft token across the primitive and composition branches, complemented by branch-specific soft prefixes for each prompt type.
% it employs a \textcolor{red}{shared primitive vocabulary with learnable soft tokens for state and object}, complemented by branch-specific soft prefixes for each prompt type. 
% \lil{To enhance cross-modal interaction, a cross-modal traction module is further introduced, using cross-attention to inject diverse visual cues into the prompt representation.} 
Given an image $x$ and a pair $(a,o)$, each branch encodes its corresponding prompt and computes temperature-scaled cosine similarities to obtain $p_a(a|x)$, $p_o(o|x)$, and $p_c((a,o)|x)$. The model is trained by minimizing the sum of cross-entropy losses across the three branches. During inference, the final composition probability is computed as a calibrated score fusion $p_c + p_a \cdot p_o$.

\subsection{Flow Matching for CZSL}
\label{sec:pre:fm}

Flow Matching~\cite{lipman2023flow,albergo2023building} learns a continuous transport between two distributions (\eg, a source $X_0$ and a target $X_1$) by regressing a time-dependent velocity field.
Rectified flow~\cite{liu2022flow} formulates a simple linear path between a sample pair $(\bm{x}_0 \sim X_0,\bm{x}_1 \sim X_1)$, which is defined as
$
\bm{x}_t = (1-t)\bm{x}_0 + t\bm{x}_1
$, where $t \in [0,1]$. The corresponding ground-truth velocity along this path is constant:
$\bm{v}^\star(\bm{x}_t, t) = \bm{x}_1 - \bm{x}_0.$
FM trains a neural network $\bm{v}_{\theta}(\bm{x}_t, t)$ to approximate this velocity field.
Concretely, at each step we sample $(\bm{x}_0, \bm{x}_1)$ and draw $t \sim \mathcal{U}[0,1]$ to construct $\bm{x}_t$, and minimize the conditional flow-matching objective:
\begin{equation}
\mathcal{L}_\mathrm{FM}(\theta)
=
\mathbb{E}_{\bm{x}_0,\bm{x}_1,t}
\bigl[
\|
\bm{v}_{\theta}(\bm{x}_t, t)
-
(\bm{x}_1 - \bm{x}_0)
\|_2^2
\bigr],
\end{equation}
which is equivalent to matching the marginal velocity field in the original FM formulation~\cite{lipman2023flow,liu2022flow}.
Once trained, samples can be transported between the two distributions by solving the ODE
$\frac{d\bm{x}_t}{dt} = \bm{v}_{\theta}(\bm{x}_t, t)$
from $t=0$ to 1 (or vice versa).
In this paper, we follow the rectified flow formulation by constructing linear interpolation paths to transport primitive visual embeddings to their corresponding textual embeddings.

% \lil{In this paper, we follow the rectified flow formulation by constructing linear interpolation paths between visual and textual embeddings for attributes and objects, respectively.}

% % \noindent \textbf{Flow Matching for CZSL.}
% FM constructs a transportation process between two distributions, $X_0$ and $X_1$, through an ordinary differential equation (ODE) defined over time $t \in [0, 1]$. 
% %
% The key idea is to learn a time-dependent velocity field $v_{\theta}(t)$, parameterized by $\theta$, that facilitates the transformation between these distributions. 
% %
% %
% Rectified flow models~\cite{liu2022flow}, a specific variant of FM, further enforce that the marginal trajectory between $X_0$ and $X_1$ follows a linear interpolation: $Z_t \sim (1 - t) X_0 + t X_1$, which allows for more efficient and stable training. 
% %
% During training, the loss is formulated as minimizing the difference between the target velocity and the predicted velocity, by solving the least squares regression problem: 
% \[
% \min_v \int_0^1 \mathbb{E} \left[ \left\| (X_1 - X_0) - v(X_t, t) \right\|^2 \right] dt,
% \]
% where $X_t$ is the linear interpolation of $X_0$ and $X_1$.
% %
% In this paper, we adopt the rectified flow framework by constructing a linear interpolation between visual and textual embeddings.
\begin{figure*}[t]
    \centering
    % 占位框
    \includegraphics[width=1\linewidth]{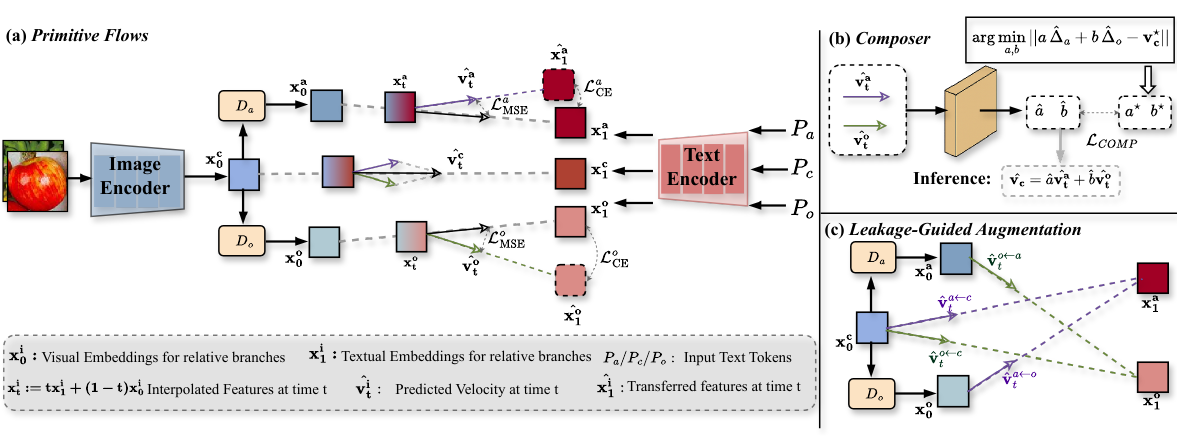}
    \caption{Overall framework for \textit{FlowComposer}. (a) Primitive flows: Two flow models learn time–conditioned velocities to transport primitive visual embeddings to their corresponding text embeddings. (b) Composer: the network to predict the combination coefficients supervised by least-squares targets. (c) Leakage-Guided Augmentation: To exploit residual cross-branch cues, each primitive flow is also trained to transport \emph{leaked} features from the counterpart (or composition) branch to its own text target.}
    \label{fig:method}
    % \vspace{-10pt}
\end{figure*}

\section{FlowComposer}
% As illustrated in Fig.~\textcolor{red}{X}, we propose \textit{FlowComposer}, a general two-path flow framework for CZSL that learns attribute and object flows separately and then composes them into a unified compositional representation in the embedding space.
% %
% Given an image $I$ with label $(a,o)$, where $a$ and $o$ denote the attribute and object label respectively, we follow the underlying CZSL baseline (\eg, CSP~\cite{nayaklearning} or Troika~\cite{huang2024troika}) with encoders $E_{\mathrm{img}}$ and $E_{\mathrm{text}}$ to obtain visual features $\bm{x}_0^{a}$, $\bm{x}_0^{o}$, $\bm{x}_0^{c}$ and textual embeddings $\bm{x}_1^{a}$, $\bm{x}_1^{o}$, $\bm{x}_1^{c}$; we keep this feature extraction intact and perform all our flow and composition operations purely within this shared feature space.

% Our framework comprises two primitive flow models, a learnable Composer, and a leakage-guided augmentation strategy. We detail the primitive flows in Sec.~\ref{sec:method:flows}, the Composer in Sec.~\ref{sec:method:composer}, and the leakage-guided augmentation in Sec.~\ref{sec:method:augmentation}.

As illustrated in Fig.~\ref{fig:method}, we propose \textit{FlowComposer}, a general flow-based framework for CZSL.
Our framework comprises two primitive flow models, a learnable Composer, and a leakage-guided augmentation strategy. We detail the primitive flows in Sec.~\ref{sec:method:flows}, the Composer in Sec.~\ref{sec:method:composer}, and the leakage-guided augmentation in Sec.~\ref{sec:method:augmentation}.

Given an image $I$ with label $(a,o)$, where $a$ and $o$ denote the attribute and object label respectively, we follow the mentioned CZSL baseline (\eg, CSP~\cite{nayaklearning} or Troika~\cite{huang2024troika}) with image encoder $E_{\mathrm{img}}$ and text encoder $E_{\mathrm{text}}$ to obtain visual features $\bm{x}_0^{a}$, $\bm{x}_0^{o}$, $\bm{x}_0^{c}$ and textual embeddings $\bm{x}_1^{a}$, $\bm{x}_1^{o}$, $\bm{x}_1^{c}$; we keep this feature extraction pipeline intact and perform all our flow and composition operations purely within this shared feature space.

% Our framework comprises two primitive flow models, a learnable Composer, and a leakage-guided augmentation strategy. We detail the primitive flows in Sec.~\ref{sec:method:flows}, the Composer in Sec.~\ref{sec:method:composer}, and the leakage-guided augmentation in Sec.~\ref{sec:method:augmentation}.

%
% Our method consists of three primary components: two independent flow models for learning transformations of attribute and object features, a learnable Composer to combine these flows effectively based on the importance of each component in a given sample, and a leakage-based augmentation strategy to leverage the existing imperfect feature disentanglement.

\subsection{Attribute and Object Flow Models}
\label{sec:method:flows}
% To capture the compositional relationships between attributes and objects, we first train two separate flow models for attribute and object, denoted as $v_{\theta_{a}}$ and $v_{\theta_{o}}$.
% %
% The attribute flow model and the object flow model each learn how to transfer corresponding visual features into their target text embeddings.
% %
% Each flow model is trained by minimizing the flow-matching \texttt{MSE} loss, which encourages the transformation of the image features to align with their corresponding text embeddings.
To support explicit composition, we first propose two separate rectified flow models, an \emph{attribute flow} $\bm{v}_{\theta_a}$ and an \emph{object flow} $\bm{v}_{\theta_o}$, which learn how to transport visual features toward their respective attribute and object text embeddings.
For each primitive branch $i \in \{a,o\}$, given the paired visual feature $\bm{x}_0^{i}$ and text embedding $\bm{x}_1^{i}$, we sample a time $t \sim \mathcal{U}[0,1]$ and construct the interpolated feature as
$\bm{x}_t^{i} = (1-t)\bm{x}_0^{i} + t \bm{x}_1^{i}$. 
The primitive flow model predicts a velocity $\hat{\bm{v}}_t^{i} = \bm{v}_{\theta_i}(\bm{x}_t^{i}, t)$, and is optimized with the flow matching MSE loss:
\begin{equation}
    \mathcal{L}_{\mathrm{MSE}}^{i}
= \mathbb{E}_{\bm{x}_0^{i},\bm{x}_1^{i},t}
\bigl[\,
\|\hat{\bm{v}}_t^{i} - (\bm{x}_1^{i} - \bm{x}_0^{i})\|_2^2
\bigr].
\end{equation}
% which trains $v_{\theta_i}$ to recover the straight-line direction from visual features to the corresponding text embeddings.
% To capture the compositional relationships between attributes and objects, we train two separate rectified-flow models, an \emph{attribute flow} $v_{\theta_a}$ and an \emph{object flow} $v_{\theta_o}$. 
% For each branch $i \in \{\mathrm{a},\mathrm{o}\}$, given the paired visual feature $\bm{x}_0^{i}$ and its target text embedding $\bm{x}_1^{i}$, we sample $t \sim \mathcal{U}[0,1]$ and construct the interpolated point
% $\bm{x}_t^{i}=(1-t)\bm{x}_0^{i}+t\bm{x}_1^{i}$. 
% The flow predicts a velocity $\hat{\bm{v}}^{\,i}_t=v_{\theta_i}(\bm{x}_t^{i},t)$ and we use a flow-matching regression loss
% \[
% \mathcal{L}_{\mathrm{MSE}}^{i}
% =\mathbb{E}_{t,\bm{x}_0^{i},\bm{x}_1^{i}}
% \Big[\;\|\hat{\bm{v}}^{\,i}_t-(\bm{x}_1^{i}-\bm{x}_0^{i})\|_2^2\;\Big],
% \]
% which trains $v_{\theta_i}$ to recover the straight-line direction from visual features to the target text embedding.
% \noindent\textbf{Endpoint classification loss.}
Besides the regression objective, we also encourage correct endpoint identification.  
We form a predicted endpoint
$
\hat{\bm{x}}_{1}^{\,i}=\bm{x}_t^{i}+(1-t)\,\hat{\bm{v}}^{\,i}_t,
$
and compute a cross-entropy loss over the vocabulary of branch $i$ by contrasting $\hat{\bm{x}}_{1}^{\,i}$ with all text embeddings $\{\bm{t}_k^{\,i}\}$ with temperature $\tau$:
\[
\mathcal{L}_{\mathrm{CE}}^{i}
=-\log
\frac{\exp\!\big(\langle \mathrm{norm}(\hat{\bm{x}}_{1}^{\,i}),\,\mathrm{norm}(\bm{x}_1^{i})\rangle/\tau\big)}
{\sum_{k}\exp\!\big(\langle \mathrm{norm}(\hat{\bm{x}}_{1}^{\,i}),\,\mathrm{norm}(\bm{t}_k^{\,i})\rangle/\tau\big)},
\]
where $\langle\cdot,\cdot\rangle$ denotes the inner product and $\mathrm{norm}(\cdot)$ denotes $l2-$normalization.
%
% \lil{Here $\langle\cdot,\cdot\rangle$ denotes xxxx and $\mathrm{norm}(\cdot)$ denotes xxxx.} 
The total loss of branch $i$ is
$
\mathcal{L}^{i}_{\mathrm{FM}}=\mathcal{L}_{\mathrm{MSE}}^{i}+\mathcal{L}_{\mathrm{CE}}^{i},
$
which jointly regresses the velocity and classifies the predicted endpoint toward the correct text target.

During inference, we adopt a one-step transport scheme for efficiency.
Given the primitive visual feature $\bm{x}_0^{i}$ and the learned flow $v_{\theta_i}$,
we approximate the endpoint embedding by evaluating the velocity at $t=0$:
$
\hat{\bm{x}}_{1}^{\,i}
= \bm{x}_0^{i} + \bm{v}_{\theta_i}(\bm{x}_0^{i}, 0),
$
which provides a direct mapping from image features to the corresponding text space without numerical integration.

\subsection{Composer}
\label{sec:method:composer}
% To capture the compositional relationships between attributes and objects, we introduce a learnable \emph{Composer} to model how they combine into a composition flow. 
% %
% For each sample, we have the primitive velocities
% $\Delta_a = x_1^a - x_0^a$, $\Delta_o = x_1^o - x_0^o$ 
% and the ground-truth composition velocity
% $\Delta_c = x_1^c - x_0^c$. 
% We first obtain target combination coefficients $(a^\star,b^\star)$ by solving a least-squares problem
% \[
% (a^\star,b^\star)
% = \arg\min_{a,b} \bigl\| a\,\Delta_a + b\,\Delta_o - \Delta_c \bigr\|_2^2,
% \]
% which specifies how the primitive flows should linearly compose to match $\Delta_c$.
% The Composer network then takes the predicted primitive velocities $(\hat{v}_a,\hat{v}_o)$ as input and outputs coefficients $(\hat{a},\hat{b})$, and is trained with
% \[
% \mathcal{L}_{\mathrm{comp}}
% = \mathbb{E}\bigl[ \|\hat{a} - a^\star\|_2^2 + \|\hat{b} - b^\star\|_2^2 \bigr],
% \]
% so that, at inference, the composition flow is given by
% $v_c = \hat{a}\,\hat{v}_a + \hat{b}\,\hat{v}_o$.
In practice, the relative contributions of attribute and object cues vary across samples. 
To capture such compositional relationships between attributes and objects, we introduce a learnable \emph{Composer} to model how the primitives combine into a composition flow in the embedding space, moving beyond token-level concatenation.
%
% Motivated by this, we aim to use \textit{Composer} to produce an explicit composition direction in the embedding space, moving beyond token-level concatenation.
% (\eg, for \textit{red apple} and \textit{ripe apple}, the contribution of attribute \textit{red} and \textit{ripe} are different )
%
% We hypothesize that the composition velocity can be expressed as a linear combination of the primitive velocities and empirically validate this assumption in \textcolor{red}{Ablation / Appendix}. 
%
For each sample, we approximate the composition velocity as a combination of primitive velocities
$
\bm{v}^\star_c = a^\star \bm{v}^\star_a + b^\star \bm{v}^\star_o,
$
and let the Composer learn the coefficients $(a^\star,b^\star)$.
For simplicity, we normalize the predicted primitive velocities to obtain unit directions
$
\hat{\Delta}_a = \frac{\hat{\bm{v}}^{\,a}_t}{\|\hat{\bm{v}}^{\,a}_t\|_2}, 
\quad
\hat{\Delta}_o = \frac{\hat{\bm{v}}^{\,o}_t}{\|\hat{\bm{v}}^{\,o}_t\|_2},
$
and define the ground-truth composition velocity as
$
\bm{v}^\star_c = \bm{x}_1^c - \bm{x}_0^c.
$
We then obtain target combination coefficients $(a^\star,b^\star)$ by solving a least-squares problem:
\[
(a^\star,b^\star)
= \arg\min_{a,b}
\bigl\|
a\,\hat{\Delta}_a + b\,\hat{\Delta}_o - \bm{v}^\star_c
\bigr\|_2^2,
\]
which specifies how the primitive directions should combine to approximate the composition velocity.
The Composer takes the predicted normalized primitive velocities
$(\hat{\bm{v}}_a, \hat{\bm{v}}_o)$ as input and outputs coefficients $(\hat{a},\hat{b})$, trained with
$
\mathcal{L}_{\mathrm{comp}}
= \mathbb{E}\bigl[
\|\hat{a} - a^\star\|_2^2 + \|\hat{b} - b^\star\|_2^2
\bigr].
$

At inference, we first predict the attribute $\hat{\bm{v}}_a$ and object velocity $\hat{\bm{v}}_o$ by primitive flows, and then compute the composition velocity
$
\hat{\bm{v}}_c = \hat{a}\,\mathrm{norm}(\hat{\bm{v}}_a) + \hat{b}\,\mathrm{norm}(\hat{\bm{v}}_o),
$
and then perform a one-step update in the embedding space
$
\hat{\bm{x}}_{1}^{c} = \bm{x}_0^{c} + h\,\hat{\bm{v}}_c,
$
where $h$ is a hyperparameter to control the step-size.

\subsection{Leakage-guided Augmentation}
\label{sec:method:augmentation}
As discussed above, visual disentanglers cannot enforce strict separation between primitives: each branch retains \emph{remained feature entanglement}, where information from one primitive inevitably leaks into the other's stream.
% Instead of treating this leakage as a flaw, we exploit it as an additional supervisory signal via \emph{leakage-guided augmentation}.
Instead of treating this leakage as a flaw, we reinterpret it as a privileged signal via \emph{leakage-guided augmentation}.

For each primitive branch $i \in \{\mathrm{a}, \mathrm{o}\}$ with target text embedding $\bm{x}_1^{i}$, we sample \emph{leaked} visual features from the other streams $j \in \{\mathrm{a}, \mathrm{o}, \mathrm{c}\}, j \neq i$, denoted as $\bm{x}_0^{j}$. 
Due to imperfect disentanglement, we posit that $\bm{x}_0^{j}$ still encodes visual cues relevant to primitive $i$. Thus, the flow $v_{\theta_i}$ should also be able to transport these leaked visual features toward $\bm{x}_1^{i}$.
Concretely, we sample $t \sim \mathcal{U}[0,1]$ and construct
$
\bm{x}_t^{i \leftarrow j} = (1-t) \bm{x}_0^{j} + t \bm{x}_1^{i},
$
then predict the velocity
$
\hat{\bm{v}}_t^{i \leftarrow j} = \bm{v}_{\theta_i}(\bm{x}_t^{i \leftarrow j}, t).
$
The leakage-guided flow-matching MSE loss is formulated as:
\[
\mathcal{L}_{\mathrm{MSE-leak}}^{i}
= \mathbb{E}_{j \neq i,\, \bm{x}_0^{j},\, \bm{x}_1^{i}, t}
\bigl[
\|\hat{\bm{v}}_t^{i \leftarrow j} - (\bm{x}_1^{i} - \bm{x}_0^{j})\|_2^2
\bigr].
\]
Similarly, we also impose a contrastive endpoint constraint on the leaked paths. 
We define the predicted endpoint
$
\hat{\bm{x}}_{1}^{\,i \leftarrow j} = \bm{x}_t^{i \leftarrow j} + (1-t) \,\hat{\bm{v}}_t^{i \leftarrow j},
$
and compute a cross-entropy loss over the vocabulary of branch $i$:
\[
\mathcal{L}_{\mathrm{CE\text{-}leak}}^{i}
= - \log
\frac{
\exp\!\left(
\langle \mathrm{norm}(\hat{\bm{x}}_{1}^{\,i \leftarrow j}),
\mathrm{norm}(\bm{x}_1^{i}) \rangle / \tau
\right)
}{
\sum_{k}
\exp\!\left(
\langle \mathrm{norm}(\hat{\bm{x}}_{1}^{\,i \leftarrow j}),
\mathrm{norm}(\bm{t}_k^{i}) \rangle / \tau
\right)
},
\]
where $\{t_k^{i}\}$ are all text embeddings of branch $i$.
The total leakage-guided loss for branch $i$ is:
\[
\mathcal{L}_{\mathrm{leak}}^{i}
= \mathcal{L}_{\mathrm{MSE\text{-}leak}}^{i}
+  \mathcal{L}_{\mathrm{CE\text{-}leak}}^{i},
\]

\begin{table*}[t]
% \vspace{-0.5em}
  \caption{Quantitative comparison (\S~\ref{sec:main_result}) on three benchmarks within closed-world and open-world setting. * denotes results from our implementation. $\ddagger$ represents the methods leveraging the LLM's knowledge.}
  \label{tab:results}
  \centering
  \setlength{\tabcolsep}{2.0pt}{
  \resizebox{1\linewidth}{!}{
\begin{tabular}{l|cccc|cccc|cccc}
\toprule
\hline
&\multicolumn{4}{c|}{MIT-States~\cite{mit}}& \multicolumn{4}{c|}{UT-Zappos~\cite{ut}}&\multicolumn{4}{c}{C-GQA~\cite{cgqa}}\\
% \cmidrule(lr){3-6} \cmidrule(lr){7-10} \cmidrule(lr){11-14} 
Method&Seen&Unseen&HM&AUC&Seen&Unseen&HM&AUC&Seen&Unseen&HM&AUC\\
\hline
\multicolumn{13}{c}{ { \it{\textbf{Closed-world} Results} } }\\
\hline 
CLIP~\cite{radford2021learning}\scriptsize \textcolor{gray}{[ICML21]}& 30.2  & 46.0  & 26.1  & 11.0  & 15.8  & 49.1  & 15.6  & 5.0   & 7.5   & 25.0  & 8.6   & 1.4 \\
CoOP~\cite{zhou2022learning}\scriptsize \textcolor{gray}{[IJCV22]} & 34.4  & 47.6  & 29.8  & 13.5  & 52.1  & 49.3  & 34.6  & 18.8  & 20.5  & 26.8  & 17.1  & 4.4 \\
DFSP(i2t)~\cite{lu2023decomposed}\scriptsize \textcolor{gray}{[CVPR23]} & 47.4  & 52.4  & 37.2  & 20.7  & 64.2  & 66.4  & 45.1  & 32.1  & 35.6  & 29.3  & 24.3  & 8.7 \\
DFSP(BiF)~\cite{lu2023decomposed}\scriptsize \textcolor{gray}{[CVPR23]} & 47.1  & 52.8  & 37.7  & 20.8  & 63.3  & 69.2  & 47.1  & 33.5  & 36.5  & 32.0  & 26.2  & 9.9 \\
DFSP(t2i)~\cite{lu2023decomposed}\scriptsize \textcolor{gray}{[CVPR23]} & 46.9  & 52.0  & 37.3  & 20.6  & 66.7  & 71.7  & 47.2  & 36.0  & 38.2  & 32.0  & 27.1  & 10.5 \\
DLM~\cite{hu2024dynamic}  \scriptsize \textcolor{gray}{[CVPR23]}   & 46.3 & 49.8 & 37.4 & 20.0   & 67.1 & 72.5 & 52.0 & 39.6  & 32.4 & 28.5 & 21.9 & 7.3  \\ 
CDS-CZSL~\cite{li2024context}  \scriptsize \textcolor{gray}{[CVPR24]}   &50.3 & 52.9 & 39.2 & 22.4    & 63.9 & 74.8 & 52.7 & 39.5    & 38.3 & 34.2 & 28.1 & 11.1     \\ 
% RAPR~\cite{jing2024retrieval} \scriptsize \textcolor{gray}{[AAAI24]}&50.0&53.3&39.2&22.5&69.4&72.8&56.5&44.5&45.6&36.0&32.0&14.4\\
IMAX\cite{10737702}                   \scriptsize \textcolor{gray}{[TPAMI25]}& 48.7& 53.8  & 39.1 & 21.9& 69.3&70.7 & 54.2 & 40.6   &39.7 & 35.8& 29.8 & 12.8      \\ 
PLID$^\ddagger$~\cite{PLID}   \scriptsize \textcolor{gray}{[ECCV24]} & 49.7 & 52.4 & 39.0 & 22.1   & 67.3 & 68.8 & 52.4 & 38.7 & 38.8 & 33.0 & 27.9 & 11.0    \\
PLO$^\ddagger$~\cite{li2023compositional} \scriptsize \textcolor{gray}{[ACMMM25]} & 51.6 & 53.7 & 40.2 & 23.4 & 70.3 & 75.8 & 55.3 & 43.6 &  44.7 & 38.1 & 33.0 & 14.9\\
LOGICZSL$^\ddagger$~\cite{wu2025logiczsl} \scriptsize \textcolor{gray}{[CVPR25]} & 50.8& 53.9& 40.5& 23.4 &69.6& 74.9 &57.8& 45.8& 44.4 &39.4 &33.3 &15.3 \\
\hline
CSP$^*$~\cite{nayaklearning}\scriptsize \textcolor{gray}{[ICLR23]} & 47.0& 49.6 & 36.6&19.6 & 63.4&65.3&47.0&32.4&24.6&29.6&19.3&5.8\\
\cellcolor{my_blue}\textbf{+FlowComposer}&\cellcolor{my_blue}48.3\upinc{3.1}&\cellcolor{my_blue}50.4\upinc{0.8}&\cellcolor{my_blue}37.6\upinc{1.0}&\cellcolor{my_blue}20.7\upinc{1.1}&\cellcolor{my_blue}66.6\upinc{3.2}&\cellcolor{my_blue}68.2\upinc{2.9}&\cellcolor{my_blue}51.2\upinc{4.2}&\cellcolor{my_blue}37.8\upinc{5.4}&\cellcolor{my_blue}29.0\upinc{4.4}&\cellcolor{my_blue}30.9\upinc{1.3}&\cellcolor{my_blue}22.9\upinc{3.6}&\cellcolor{my_blue}7.7\upinc{1.9} \\
% \hline
Troika$^*$~\cite{huang2024troika} \scriptsize \textcolor{gray}{[CVPR24]}& 49.3&52.5&39.2&22.1& 66.3&73.4&55.4&41.8&41.0&35.7&29.7&12.5 \\
\cellcolor{my_blue}\textbf{+FlowComposer}& \cellcolor{my_blue}51.5\upinc{2.2}&\cellcolor{my_blue}53.2\upinc{0.7}&\cellcolor{my_blue}40.2\upinc{1.0}&\cellcolor{my_blue}23.5\upinc{1.4}& \cellcolor{my_blue}71.1\upinc{4.8}&\cellcolor{my_blue}74.9\upinc{1.5}&\cellcolor{my_blue}58.6\upinc{3.2}&\cellcolor{my_blue}46.8\upinc{5.0}&\cellcolor{my_blue}44.8\upinc{3.8}&\cellcolor{my_blue}40.7\upinc{5.0}&\cellcolor{my_blue}34.0\upinc{4.3}&\cellcolor{my_blue}15.9\upinc{3.4} \\
\hline
\multicolumn{13}{c}{ { \it{\textbf{Open-world} Results} } }\\
\hline
CLIP~\cite{radford2021learning}\scriptsize \textcolor{gray}{[ICML21]}& 30.1  & 14.3  & 12.8  & 3.0   & 15.7  & 20.6  & 11.2  & 2.2   & 7.5   & 4.6   & 4.0   & 0.27 \\
CoOP~\cite{zhou2022learning}\scriptsize \textcolor{gray}{[IJCV22]} & 34.6  & 9.3   & 12.3  & 2.8   & 52.1  & 31.5  & 28.9  & 13.2  & 21.0  & 4.6   & 5.5   & 0.70 \\
DFSP(i2t)~\cite{lu2023decomposed}\scriptsize \textcolor{gray}{[CVPR23]}& 47.2  & 18.2  & 19.1  & 6.7   & 64.3  & 53.8  & 41.2  & 26.4  & 35.6  & 6.5   & 9.0   & 1.95 \\
DFSP(BiF)~\cite{lu2023decomposed}\scriptsize \textcolor{gray}{[CVPR23]} & 47.1  & 18.1  & 19.2  & 6.7   & 63.5  & 57.2  & 42.7  & 27.6  & 36.4  & 7.6   & 10.6  & 2.39 \\
DFSP(t2i)~\cite{lu2023decomposed}\scriptsize \textcolor{gray}{[CVPR23]} & 47.5  & 18.5  & 19.3  & 6.8   & 66.8 & 60.0  & 44.0  & 30.3  & 38.3  & 7.2   & 10.4  & 2.40  \\
CDS-CZSL~\cite{li2024context} \scriptsize \textcolor{gray}{[CVPR24]} & 49.4 & 21.8 & 22.1 & 8.5 & 64.7 & 61.3 & 48.2 & 32.3 & 37.6 & 8.2 & 11.6 & 2.7   \\
% RAPR~\cite{jing2024retrieval} \scriptsize \textcolor{gray}{[AAAI24]}&50.0&53.3&39.2&22.5&69.4&72.8&56.5&44.5&45.6&36.0&32.0&14.4\\
IMAX\cite{10737702}                \scriptsize \textcolor{gray}{[TPAMI25]} & 50.2& 18.6  & 21.4 & 7.6 & 68.4&57.3 & 47.5 & 32.3   &38.7 & 7.9& 11.2 & 2.5    \\ 
PLID$^\ddagger$~\cite{PLID}   \scriptsize\textcolor{gray}{[ECCV24]}& 49.1 & 18.7 & 20.4 & 7.3& 67.6 & 55.5 & 46.6 & 30.8 &39.1 & 7.5 & 10.6 & 2.5 \\   
PLO$^\ddagger$~\cite{li2023compositional}\scriptsize\textcolor{gray}{[ACMMM25]} & 49.7 & 19.4 & 21.4 & 7.8 & 68.0 & 63.5 & 47.8 & 33.1 & 43.9 & 10.4 & 13.9 & 3.9 \\
LOGICZSL$^\ddagger$~\cite{wu2025logiczsl} \scriptsize \textcolor{gray}{[CVPR25]} & 50.7 &21.4& 22.4& 8.7 &69.6& 63.7 &50.8 &36.2& 43.7& 9.3& 12.6 &3.4\\
\hline
CSP$^*$~\cite{nayaklearning}\scriptsize\textcolor{gray}{[ICLR23]} &  47.4&15.3&17.1&5.6
 &  63.4&44.8&40.6&23.3&29.6&5.4&7.3&1.3\\
\cellcolor{my_blue}\textbf{+FlowComposer}&\cellcolor{my_blue}48.2\upinc{0.8}&\cellcolor{my_blue}15.6\upinc{0.3}&\cellcolor{my_blue}17.8\upinc{0.7}&\cellcolor{my_blue}6.1\upinc{0.5}&\cellcolor{my_blue}66.4\upinc{3.0}&\cellcolor{my_blue}46.2\upinc{1.4}&\cellcolor{my_blue}41.2\upinc{0.6}&\cellcolor{my_blue}25.3\upinc{2.0}&\cellcolor{my_blue}31.2\upinc{0.6}&\cellcolor{my_blue}5.8\upinc{0.4}&\cellcolor{my_blue}8.0\upinc{0.7}&\cellcolor{my_blue}1.8\upinc{0.5} \\
\hline
Troika$^*$~\cite{huang2024troika}\scriptsize \textcolor{gray}{[CVPR24]}& 50.3&17.5&19.0&6.8& 65.8&61.0&46.6&32.7& 40.8  &8.6 &11.7 &2.9 \\
\cellcolor{my_blue}\textbf{+FlowComposer}& \cellcolor{my_blue}50.4\upinc{0.1}&\cellcolor{my_blue}19.0\upinc{1.5}&\cellcolor{my_blue}20.3\upinc{1.3}&\cellcolor{my_blue}7.5\upinc{0.7}& \cellcolor{my_blue}70.1\upinc{4.3}&\cellcolor{my_blue}61.2\upinc{0.2}&\cellcolor{my_blue}51.0\upinc{4.4}&\cellcolor{my_blue}35.5\upinc{2.8} &\cellcolor{my_blue}43.5 \upinc{2.7}&\cellcolor{my_blue} 10.2\upinc{1.6} & \cellcolor{my_blue}12.6\upinc{0.9}&\cellcolor{my_blue}3.5\upinc{0.6}\\
\bottomrule
\end{tabular}
}
}
\vspace{-10pt}
% \raggedright\footnotesize * Results from our implementation.
\end{table*}

\section{Experiment}

\subsection{Setups and Implementations}
\noindent\textbf{Datasets.}
We conduct a comprehensive evaluation of our method across three real-world CZSL benchmarks: (i) MIT-States~\cite{mit} (ii) UT-Zappos~\cite{ut} (iii) C-GQA~\cite{cgqa}. We follow the data split protocol of \cite{nayaklearning,purushwalkam2019task} to partition datasets into train, validation, and test datasets. 

\noindent\textbf{Evaluation Metrics.}
Following the standard CZSL protocol~\cite{nayaklearning,huang2024troika}, we apply a calibration bias to the \emph{unseen} scores and sweep it from $-\infty$ to $+\infty$ to balance predictions between seen and unseen compositions. 
Over this sweep, we report the Area Under the Curve (\textbf{AUC}) of the seen–unseen trade-off and the best harmonic mean (\textbf{HM}). 
For completeness, we also record the best-seen accuracy \textbf{Seen} and the best-unseen accuracy \textbf{Unseen} achieved during the sweep.
For open-world evaluation, we adopt the post-training calibration strategy of \cite{nayaklearning} to discard infeasible compositions.

\noindent\textbf{Implementation Details.}
We follow standard CZSL settings and use the pre-trained CLIP ViT-L/14~\cite{radford2021learning} as both image and text encoder. 
We implement \textit{FlowComposer} on two popular CZSL baselines - CSP~\cite{nayaklearning} and Troika~\cite{huang2024troika}.
For the single-path baseline~\cite{nayaklearning}, we derive attribute and object text embeddings using simple prompts (``\textit{a photo of} \texttt{[attribute]}'' / ``\textit{a photo of} \texttt{[object]}''), and apply our primitive flows and composer accordingly.
The leakage-guided augmentation is used only with the multi-path methods~\cite{huang2024troika}, since single-path method~\cite{nayaklearning} does not provide branch-wise visual disentanglement.
We employ a lightweight architecture proposed in MAR~\cite{li2024autoregressive} as flow matching network, implemented as a deep residual MLP with timestep conditioning. 
For \textit{Composer}, we employ a small three-layer MLP with LayerNorm and GELU~\cite{hendrycks2023gaussianerrorlinearunits} activations.
% Compared to generative models that operate in high-dimensional pixel or latent spaces, our flow operates on much lower-dimensional feature vectors, so a compact network is sufficient to model the velocity field. This choice avoids introducing heavy U-Net–style backbones and keeps the additional computational and memory overhead of our method negligible while still providing enough capacity to learn accurate flows.
%
% All models are trained and evaluated on a single NVIDIA RTX \textcolor{red}{3090 or A800} GPU.
%
Further details on datasets, optimization, and hyperparameters are deferred to the supplementary material.

% \input{tabs/abl_comp+flows+step}
% 导言区：
% \usepackage{pifont}
% \newcommand{\cmark}{\ding{51}} % 对勾
% \newcommand{\xmark}{\ding{55}} % 叉

% \begin{table*}[ht]
% \small
% \centering
% \caption{Ablations. The results regarding the different components in our \textit{FlowComposer} on MIT-States~\cite{mit} and UT-Zappos~\cite{ut}.}
% \resizebox{0.7\linewidth}{!}{
% \begin{tabular}{ccc|cccc|cccc}
% \toprule \hline
% \textit{Flows} &\textit{Composer} & \textit{LG-Aug} & \multicolumn{4}{c|}{MIT-States~\cite{mit}} & \multicolumn{4}{c}{UT-Zappos~\cite{ut}} \\
% (\S~\ref{sec:method:flows}) & (\S~\ref{sec:method:composer}) & (\S~\ref{sec:method:augmentation}) & Seen & Unseen & HM & AUC & Seen & Unseen & HM & AUC \\
% \midrule
% \xmark & \xmark & \xmark & 49.3&52.5&39.2&22.1& 66.3&73.4&55.4&41.8 \\
% \cmark & \xmark & \xmark & 51.3 & 52.1 & 39.4 & 22.6 & 68.8 & 74.8 & 55.1 &43.0  \\
% \cmark & \xmark & \cmark & 51.3 & 53.0 & 40.3 & 23.3 & 69.2 & 74.9 &56.3  &44.0  \\
% \cmark & \cmark & \xmark & 51.0 & 53.0 & 40.1 & 23.3 & 69.7 & 74.9 & 58.5 & 45.8 \\
% \cmark & \cmark & \cmark & 51.5 & 53.2 & 40.2 & 23.5 & 71.1 & 74.9 & 58.6 & 46.8 \\
% \bottomrule
% \end{tabular}}
% \label{tab:ablation}
% \end{table*}

\begin{table*}[ht]
\small
\centering
\vspace{-8pt}
\caption{Ablations. The results regarding the different components in our \textit{FlowComposer} on MIT-States~\cite{mit} and UT-Zappos~\cite{ut}.}
\setlength\tabcolsep{5.0pt}
\resizebox{0.8\linewidth}{!}{
\begin{tabular}{cccc|cccc|cccc}
\toprule \hline
&\textit{Flows} &\textit{Composer} & \textit{LG-Aug} & \multicolumn{4}{c|}{MIT-States~\cite{mit}} & \multicolumn{4}{c}{UT-Zappos~\cite{ut}} \\
&(\S~\ref{sec:method:flows}) & (\S~\ref{sec:method:composer}) & (\S~\ref{sec:method:augmentation}) & Seen & Unseen & HM & AUC & Seen & Unseen & HM & AUC \\
\midrule
Baseline&\xmark & \xmark & \xmark & 49.3&52.5&39.2&22.1& 66.3&73.4&55.4&41.8 \\
(1)&\cmark & \xmark & \xmark & 51.3 & 52.1 & 39.4 & 22.6 & 68.8 & 74.8 & 55.1 &43.0  \\
(2)&\cmark & \xmark & \cmark & 51.3 & 53.0 & \textbf{40.3} & 23.3 & 69.2 & 74.9 &56.3  &44.0  \\
(3)&\cmark & \cmark & \xmark & 51.0 & 53.0 & 40.1 & 23.3 & 69.7 & 74.9 & 58.5 & 45.8 \\
\rowcolor{my_blue}Ours&\cmark & \cmark & \cmark & \textbf{51.5} & \textbf{53.2} & 40.2 & \textbf{23.5} & \textbf{71.1} & \textbf{74.9} & \textbf{58.6} & \textbf{46.8} \\
\bottomrule
\end{tabular}}
\label{tab:ablation}
\vspace{-12pt}
\end{table*}

% \begin{table}[ht]
% \small
% \centering
% \caption{Ablation.}
% \resizebox{\linewidth}{!}{
% \begin{tabular}{ccc|cccc|cccc}
% \toprule \hline
%  & & & \multicolumn{4}{c|}{MIT-States~\cite{mit}} & \multicolumn{4}{c}{UT-Zappos~\cite{ut}} \\
% Flows & Composer & Aug & Seen & Unseen & HM & AUC & Seen & Unseen & HM & AUC \\
% \midrule
% \xmark & \xmark & \xmark &  &  &  &  &  &  &  &  \\
% \cmark & \xmark & \xmark &  &  &  &  &  &  &  &  \\
% \cmark & \xmark & \cmark &  &  &  &  &  &  &  &  \\
% \cmark & \cmark & \xmark &  &  &  &  &  &  &  &  \\
% \cmark & \cmark & \cmark &  &  &  &  &  &  &  &  \\
% \bottomrule
% \end{tabular}}
% \label{tab:ablation}
% \end{table}

\subsection{Main Results}
\label{sec:main_result}
Tab.~\ref{tab:results} reports quantitative comparisons with recent CZSL methods on three benchmarks (MIT-States~\cite{mit}, UT-Zappos~\cite{ut} and C-GQA~\cite{cgqa}) under both closed-world and open-world settings.
Plugging our \textit{FlowComposer} into both CSP~\cite{nayaklearning} and Troika~\cite{huang2024troika} consistently improve the performance in four metrics across all datasets and settings, showing that our flow-based formulation is robust and model-agnostic.
In the \textbf{closed-world} setting, \textit{FlowComposer} on Troika~\cite{huang2024troika} achieves state-of-the-art AUC on all three datasets, even surpassing approaches that exploit additional LLM-generated knowledge (\eg, PLID~\cite{PLID}, PLO~\cite{li2023compositional}, LOGICZSL~\cite{wu2025logiczsl}).
Specifically, we improve AUC over Troika~\cite{huang2024troika} by $+1.4\%$ on MIT-States, $+5.0\%$ on UT-Zappos and $+3.4\%$ on C-GQA.
In the \textbf{open-world} setting, \textit{FlowComposer} brings substantial HM and AUC gains over CSP~\cite{nayaklearning} and Troika~\cite{huang2024troika}, \eg, $+1.3\%$ HM on MIT-States~\cite{mit}, $+4.4\%$ HM on UT-Zappos~\cite{ut}, and $+0.9\%$ on C-GQA~\cite{cgqa} compared with Troika.
The consistent improvements over baselines for all datasets and settings illustrate the effectiveness and robustness of our \textit{FlowComposer}, and highlight flow matching as a natural and well-suited paradigm for CZSL.
Although our method is slightly behind LOGICZSL~\cite{wu2025logiczsl} on AUC under the open-world setting, we note that LOGICZSL benefits from extra logic rules derived from LLMs, whereas our method attains strong generalization without using external textual knowledge.

% We conduct ablation studies to analyse the contributions of each major component in our framework: Hierarchical Learning, Consistency Self-Distillation, and Hierarchical Semantic-Guided Soft Contrastive Learning (HSCL). As shown in Tab.~\ref{tab:ablation}, we report results on the SSB benchmark~\cite{vaze2022semantic}, including Stanford Cars~\cite{krause20133d}, CUB~\cite{wah2011caltech}, and FGVC-Aircraft~\cite{maji2013fine} datasets, evaluated over `All', `Old', and `New' categories.
% %
% Starting from the baseline trained solely with the GCD loss, we incrementally integrate the proposed components. Incorporating hierarchical learning alone (Row (1)) yields a modest improvement, particularly on the old categories. Adding consistency-based self-distillation (Row (2)) further improves alignment and stability, while semantic-guided HSCL (Row (3)) significantly boosts performance on novel classes by leveraging cross-instance semantic similarity. When all components are combined, the full framework achieves substantial gains with $11.5\%$ on Stanford Cars, $7.8\%$ on FGVC-Aircraft, and $5.7\%$ on CUB.

\subsection{Diagnostic Analysis}

% The single-flow baseline ignores primitive structure and tends to couple attribute–object interactions into a single trajectory, which we found hurts transfer to unseen pairs (lower HM/AUC). 
% Conversely, training an additional composition flow introduces a shortcut that bypasses primitive reasoning and increases computation, yielding marginal accuracy gains and weaker open-world generalization.
% Our \emph{two-flow} design (attribute \& object) with the \emph{Composer} consistently strikes the best balance: it preserves primitive factorization, composes velocities explicitly in the embedding space, and improves HM/AUC over the single-flow while matching or surpassing the three-flow variant with fewer parameters and negligible inference overhead.

\noindent\textbf{Component Analysis.} 
We conduct ablations to quantify the contribution of each module: \emph{Primitive Flows (Flows)}, \emph{Composer}, and \emph{Leakage-Guided Augmentation (LG-Aug)}.
As shown in Tab.~\ref{tab:ablation}, we report results under the closed-world setting on MIT-States~\cite{mit} and UT-Zappos~\cite{ut}, comparing against the Troika~\cite{huang2024troika} baseline.
Starting from the baseline, we incrementally integrate the proposed components. 
Four observations can be drawn. \textbf{First}, incorporating \textit{primitive flows} (Row (1)) yields a modest improvement particularly on the seen compositions.
\textbf{Second}, adding \textit{Leakage-Guided Augmentation} (Row (2)) further improves alignment and robustness, leading to improvements in AUC and HM ($+0.9\%$ HM on MIT-States and $+1.2\%$ HM on UT-Zappos).
\textbf{Third}, alternatively, introducing the \emph{Composer} on top of flows (Row (3)) also stabilizes recognition, increasing both AUC and HM.
\textbf{Fourth}, combining all components, the full framework achieves substantial gains over all metrics, indicating that explicit velocity composition and leakage-guided supervision are complementary to flow learning.

\noindent\textbf{Primitive Flows \textit{vs.} Composition Flows.}
% \noindent\textbf{Number of flows.}
To justify the design choice of using two primitive flows, we ablate (i) a \emph{single-flow} variant that learns one cross-modal flow directly for composition ($c$ only), and (ii) a \emph{three-flow} variant that adds an extra composition flow on top of the attribute/object flows (\textit{c+a+o}) to replace the composer (see first four rows of Tab.~\ref{tab:flows+h}). 
As seen, whether using a single composition flow (\textit{c}) or a three-flow setup (\textit{c+a+o}), unseen-composition accuracy degrades (around $4\%$ deduction for MIT-States~\cite{mit} and over $10\%$ deduction for UT-Zappos~\cite{ut}).
We attribute this to a lack of open-world capacity: a composition flow directly models seen pair trajectories, but unseen compositions never appear in training, so the model overfits to seen pair geometry.
%
% In contrast, our primitive flows aim to learn attribute and object transports where all those attribute and object categories are seen during training. composing these transports at test time confers better generalization to novel attribute–object combinations.
%
In contrast, primitive flows learn attribute and object transports where all those attribute and object categories are seen during training; composing these transports at test time confers better generalization to unseen attribute–object combinations.

\begin{table}[ht]
\centering
% \vspace{-5pt}
\caption{Analysis on flows (\S~\ref{sec:method:flows}) on the MIT-States~\cite{mit} and UT-Zappos~\cite{ut} under closed-world setting. }
\setlength\tabcolsep{3.5pt}
\resizebox{1.0\linewidth}{!}{
\begin{tabular}{ccccccccc}
\toprule 
&\multicolumn{4}{c}{MIT-States}&\multicolumn{4}{c}{UT-Zappos}\\
\cmidrule(lr){2-5} \cmidrule(lr){6-9}
  & S & U & H & A & S & U & H & A \\
\midrule
% \multirow{3}{*}{\rotatebox[origin=c]{90}{\textit{MIT}}}
\textit{N/A}&49.3&52.5&39.2&22.1& 66.3&73.4&55.4&41.8 \\
\textit{c } &50.9&48.0&36.9&20.0&70.0&60.1&44.6&37.2 \\
\textit{c+a+o}&51.0&48.7&37.4&20.5& 70.1&60.6&44.9&37.6\\
\rowcolor{my_blue}\textit{FlowComposer}& \textbf{51.5} & \textbf{53.2} & \textbf{40.2} & \textbf{23.5} & \textbf{71.1} & \textbf{74.9} & \textbf{58.6} & \textbf{46.8}\\
% \hline
% % ResMLP\\
% % Flows&51.3 & 52.1 & 39.4 & 22.6 & 68.8 & 74.8 & 55.1 &43.0  \\
% % \hline
% \textbf{$h=0.1$}  & \textbf{51.5} & \textbf{53.2} & \textbf{40.2} & \textbf{23.5} & 69.6 & 71.8 & 56.5 &43.8 \\
%   $h=0.5 $ & 51.0 & 52.5 & 39.7 & 22.9& 70.1 & 72.9 & 57.4 &45.0  \\
% $h=1.0$  & 49.5 & 49.6 & 37.5 & 20.6 & \textbf{71.1} & \textbf{74.9} & \textbf{58.6} & \textbf{46.8} \\
\bottomrule
\end{tabular}
}
\vspace{-8pt}
\label{tab:flows+h}
\end{table}

\begin{figure*}[ht]
    \centering
    \includegraphics[width=1\textwidth]{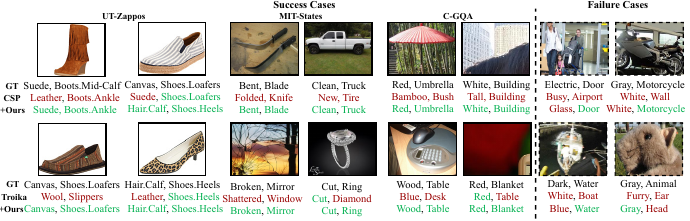}
    \caption{Visual comparisons between CSP~\cite{nayaklearning}, Troika~\cite{huang2024troika} and our \textit{FlowComposer} on three datasets. \textcolor[HTML]{990000}{Red} represents the wrong prediction, and \textcolor[HTML]{02B150}{Green} represents the right prediction.
    }
    \label{fig:vis}
    \vspace{-12pt}
\end{figure*}

\noindent\textbf{Visualization.}
Fig.~\ref{fig:vis} shows top-1 predictions between  two baselines ( CSP~\cite{nayaklearning} and Troika~\cite{huang2024troika}) and our \textit{FlowComposer}.
For success cases, \textit{FlowComposer} is more reliable than baseline methods, correctly identifying compositions that the baseline misclassifies.
For instance, it predicts \texttt{Wood Table} even when key visual cues are occluded, where Troika fails, and it further corrects both attribute and object for some cases (\eg, from \texttt{Leather} to \texttt{Hair.Calf}, from \texttt{Table} to \texttt{Blanket}, from \texttt{New Tire} to \texttt{Clean Truck}), reflecting stronger sensitivity to fine-grained primitives.
For failure cases, our predictions remain semantically close to the ground truth, such as \texttt{White Motorcycle} for \texttt{Gray Motorcycle} or \texttt{Blue Water} for \texttt{Dark Water}, indicating that even when incorrect, \textit{FlowComposer} tends to produce plausible compositions rather than irrelevant ones.

\noindent\textbf{Composer \textit{vs.} Predictor.}
To further analyze the rationale of the \emph{Composer}, we ablate it with a \emph{direct predictor} variant that takes the same primitive velocities as inputs and with the same model structure but regresses the composition velocity outright, instead of predicting combination coefficients. 
We report the comparison between predictor and composer in Tab.~\ref{tab:linear} based on the Troika+\textit{FlowComposer} setup.
The predictor can be treated as a non-linear counterpart that maps primitive velocities to the composition velocity.
This non-linear variant consistently underperforms our Composer (\eg, $0.2\%$ AUC deduction in MIT-States~\cite{mit}, $1.6\%$ AUC deduction in UT-Zappos~\cite{ut} and $0.2\%$ AUC deduction in C-GQA~\cite{cgqa}), supporting the design choice of learning an explicit combination rule.

% 需要在导言区
% \usepackage{graphicx}
% \usepackage{multirow}

\begin{table}[ht]
\centering
% \vspace{-5pt}
\caption{Comparison between Predictor and Composer in MIT-States~\cite{mit}, UT-Zappos~\cite{ut} and C-GQA~\cite{cgqa} on HM and AUC under the closed-world setting.}
\setlength{\tabcolsep}{8pt}
\resizebox{1.0\linewidth}{!}{
\begin{tabular}{lcccccc}
\toprule 
&\multicolumn{2}{c}{MIT-States}&\multicolumn{2}{c}{UT-Zappos}&\multicolumn{2}{c}{C-GQA}\\
\cmidrule(lr){2-3} \cmidrule(lr){4-5} \cmidrule(lr){6-7}
 & HM & AUC & HM & AUC & HM & AUC \\
\midrule
Predictor& 39.9 & 23.3 & 58.6 & 45.2 & 33.9 & 15.7  \\
\rowcolor{my_blue}\textit{\textbf{Composer}}& 40.2 & 23.5 & 58.6 & 46.8 & 34.0 & 15.9  \\
\hline
\end{tabular}
}
\label{tab:linear}
\vspace{-2pt}
\end{table}

% \subsection{FM \textit{vs} PEFT}
\noindent\textbf{FM \textit{vs} Regressor.}
To verify that our improvements stem from the \emph{flow-matching formulation} rather than extra parameters, we add a parameter-matched regressor baseline. Concretely, we reuse the \emph{same} network architecture as our velocity model, but use it to predict an endpoint residual in feature space: $\hat{x}_1^{\,i}=x_0^{i}+\Delta_{\phi}(x_0^{i})$, where $\Delta_{\phi}$ is trained to minimize a classification loss against the target text embedding. This removes the time-conditioning and velocity supervision, keeping total trainable parameters and FLOPs comparable to our FM branch.
As reported in Tab.~\ref{tab:supp:adapter}, the parameter-matched regressor lags behind our flow matching based approach across all three datasets and metrics, despite equal capacity and similar inference cost. These results indicate that the gain comes from our design of using flow matching, not from adding parameters or larger adapters.
Taken together, this comparison suggests that time-conditioned velocity supervision is inherently more suitable for capturing composition dynamics than merely attaching a residual regression head, even when the number of trainable parameters is matched.
\begin{table}[h]
\centering
% \vspace{-5pt}
\caption{Comparison between regressor and our \textit{FM} on HM and AUC under the closed-world setting.}
\setlength{\tabcolsep}{8pt}
\resizebox{1.0\linewidth}{!}{
\begin{tabular}{lcccccc}
\toprule 
&\multicolumn{2}{c}{MIT-States}&\multicolumn{2}{c}{UT-Zappos}&\multicolumn{2}{c}{C-GQA}\\
\cmidrule(lr){2-3} \cmidrule(lr){4-5} \cmidrule(lr){6-7}
 & HM & AUC & HM & AUC & HM & AUC \\
\midrule
Regressor&39.6 &22.5 & 55.2&42.8&30.0&12.9 \\
\rowcolor{my_blue}\textit{\textbf{FM}}& 40.2 & 23.5 & 58.6 & 46.8 & 34.0 & 15.9  \\
\hline
\end{tabular}
}
\label{tab:supp:adapter}
\vspace{-12pt}
\end{table}

\section{Conclusion}
\label{sec:conclusion}
In this paper, we take a first step towards leveraging flow matching for compositional zero-shot learning, motivated by the natural compositionality and decompositionality of velocity fields in the embedding space.
\textit{First}, we introduce \textit{FlowComposer}, a general framework that learns attribute and object flows and explicitly composes them via a learnable Composer, replacing implicit \textit{token-level} concatenation with an operational composition rule in the embedding space.
\textit{Second}, we propose a leakage-guided augmentation strategy that converts cross-branch leakage into useful supervisory signals to further exploit inevitable residual entanglement in existing CZSL pipelines.
\textit{Third}, we conduct extensive experiments on three public CZSL benchmarks and two strong VLM-based baselines which demonstrate that \textit{FlowComposer} consistently improves HM and AUC, yields a more balanced trade-off between seen and unseen performance, and incurs only negligible inference overhead, further supporting flow matching as a natural and effective paradigm for CZSL.

\paragraph{Acknowledgments.}
This work was supported by the National Natural Science Foundation of China Young Scholar Fund Category C (62402408), the Hong Kong SAR RGC Early Career Scheme (26208924), the National Natural Science Foundation of China Young Scholar Fund Category B (62522216), and the Hong Kong SAR RGC General Research Fund (16219025).
Thanks to Yanghao Wang for GPU support.

{
    \small
    \bibliographystyle{ieeenat_fullname}
    \bibliography{main}
}
\newpage
\maketitlesupplementary
We provide additional details and experimental results in this supplementary material, which is organized as follows:
\begin{itemize}
\item \S\ref{sec:details} More Experimental Details
\item \S\ref{sec:supp:remained} Remained Feature Entanglement Analysis
\item \S\ref{sec:supp:analysis} Additional Analysis
\item \S\ref{sec:supp:5runs} Stability Analysis
\item \S\ref{sec:supp:composer} Semantic Analysis of Composer
\item \S\ref{sec:supp:impact} Social Impacts
\end{itemize}

\section{More Experimental Details}
\label{sec:details}
\subsection{Dataset Statistics}
For each dataset, we adhere to the data splitting scheme from \cite{nayaklearning,purushwalkam2019task}. Detailed data statistics are reported in Tab.~\ref{tab:dataset_stats}. MIT-States~\cite{mit} is a large-scale attribute--object benchmark containing diverse everyday objects paired with a wide range of state attributes (\eg, \textit{sliced}, \textit{moldy}, \textit{wet}), resulting in a dense and visually heterogeneous composition space. UT-Zappos~\cite{ut} focuses on fine-grained footwear recognition, where attributes primarily describe material or surface appearance (\eg, \textit{leather}, \textit{satin}, \textit{suede}) and objects correspond to specific shoe categories. C-GQA~\cite{cgqa} extends the GQA dataset with compositional attribute--object annotations, covering a much broader variety of scenes, colors, textures, and object types, and thus presents a more diverse and challenging open-world composition landscape.

\begin{table}[h]
  \centering
\caption{Statistics of three datasets in our experiments. 
  The number of elements in each set is reported.
  }
  \resizebox{\linewidth}{!}{
    \begin{tabular}{lcc|cc|ccc|ccc}
    \toprule
          &       &       & \multicolumn{2}{c|}{\textbf{Training}} & \multicolumn{3}{c|}{\textbf{Validation}} & \multicolumn{3}{c}{\textbf{Test}} \\
          \hline
    \textbf{Dataset} & $\mathcal{S}$ & $\mathcal{O}$ & $\mathcal{C}^{se}$ & $\mathcal{X}$ & $\mathcal{C}^{se}$ & $\mathcal{C}^{us}$ & $\mathcal{X}$ & $\mathcal{C}^{se}$ & $\mathcal{C}^{us}$ & $\mathcal{X}$ \\
    \hline
    MIT-States~\cite{mit} & 115   & 245   & 1262  & 30k   & 300   & 300   & 10k   & 400   & 400   & 13k \\
    UT-Zappos~\cite{ut} & 16    & 12    & 83    & 23k   & 15    & 15    & 3k    & 18    & 18    & 3k \\
    C-GQA~\cite{cgqa} & 413     & 674     & 5592    & 27k    & 1252    & 1040    & 7k    & 888     & 923     & 5k \\
    \bottomrule
    \end{tabular}
    }
  \label{tab:dataset_stats}
\end{table}

\subsection{Additional Implementation Details}
% machine / optimizer / structure
Consistent with prior studies~\cite{nayaklearning,huang2024troika}, we adopt the official OpenAI checkpoints of CLIP with a ViT-L/14~\cite{radford2021learning} configuration as our visual–textual backbone. For data preprocessing and training configuration, we follow the same setup as Troika~\cite{huang2024troika} and CSP~\cite{nayaklearning}: images are processed with the same resizing and cropping pipeline, and Adam is used as the optimizer for all datasets.
For \textit{FlowComposer}, we utilize AdamW~\cite{loshchilov2017decoupled} as the optimizer for both primitive flows and the composer.

\noindent\textbf{Flow Matching Networks.} As introduced in the main paper Sec.~5.1,  We employ a lightweight architecture proposed in MAR~\cite{li2024autoregressive} as flow matching network, implemented as a deep residual MLP with timestep conditioning. 
Concretely, the input feature $\bm{x}\in\mathbb{R}^{D}$ (CLIP feature of dimension $D$) is fed into a stack of 24 ResBlocks with adaptive LayerNorm (adaLN) modulation, where timestep embeddings from a small MLP (\textit{TimestepEmbedder}) control per-channel shift, scale, and gating, and each block uses SiLU activations.

\noindent\textbf{Composer Networks.} We implement the \textit{Composer} as a lightweight residual MLP that takes the attribute and object velocities $\bm{v}_a,\bm{v}_o \in \mathbb{R}^D$ as input, and processes them with LayerNorm, GELU activations, and a small stack of residual MLP blocks before projecting back to $\mathbb{R}^D$. 
We report the details of the model parameters in Tab.~\ref{tab:supp:param}.

\begin{table}[h]
\centering
\caption{Model size and computational cost of our networks.}
\begin{tabular}{lcc}
\toprule
Network & \# Params (M)$\downarrow$ & GFLOPs$\downarrow$ \\
\midrule
FM        & 74.06 & 0.07 \\
Composer  & 25.97 & 0.03 \\
\bottomrule
\end{tabular}
\label{tab:supp:param}
\end{table}

\section{Remained Feature Entanglement Analysis}
\label{sec:supp:remained}
\subsection{Leakage Analysis}
To further support our argument about \textit{Remained Feature Entanglement}, we quantitatively probe the amount of cross-branch “leakage’’ in the multi-path baseline Troika~\cite{huang2024troika}.
Our hypothesis is that, under perfect disentanglement, the attribute branch should encode only attribute-related information and be essentially uninformative for object prediction (and vice versa).
In that case, using attribute features to classify objects (or object features to classify attributes) should yield accuracies close to random chance.
To test this, we extract the attribute, object, and composition features from Troika~\cite{huang2024troika} on the test sets and use each branch to predict both attribute and object labels with the same CLIP-style cosine classifier aligned with Troika's inference. We then measure the resulting attribute and object accuracies for all three branches on all three datasets, and report the results in Tab.~\ref{tab:supp:leakage}. 

In Tab.~\ref{tab:supp:leakage} we report \emph{class-balanced accuracy} (per-class average).
The first row (Random) is a chance baseline with randomly generated predictions.
The second row (\textit{Attr Branch}) uses the \emph{attribute} image features to classify both attributes and objects via the same CLIP-style cosine classifier; the third row (\textit{Obj Branch}) does the symmetric test with \emph{object} features; the fourth row (\textit{Comp Branch}) uses the \emph{composition} features.
Clear leakage emerges: using attribute features to predict \emph{objects} is far above chance (MIT: \textbf{3.87} vs.\ 0.47, UT: \textbf{14.9} vs.\ 8.14, C-GQA: \textbf{2.36} vs.\ 0.05).
Conversely, using object features to predict \emph{attributes} also surpasses chance (MIT: \textbf{6.3} vs.\ 1.13, UT: \textbf{7.69} vs.\ 5.74, C-GQA: \textbf{2.12} vs.\ 0.25).
Moreover, composition features are strongly predictive for both primitives (\eg, object accuracy: MIT \textbf{51.3}, UT \textbf{58.9}, C-GQA \textbf{39.8}), indicating substantial cross-branch carry-over.
These findings are inconsistent with perfect disentanglement and quantitatively support our claim of \textit{remained feature entanglement}.

\begin{table}[h]
\centering
% \vspace{-5pt}
\caption{Comparison between random prediction and predictions from \textit{Attribute}, \textit{Object} and \textit{Composition} branches on class-balanced accuracy (\textit{Acc}) for attributes and objects.}
\setlength{\tabcolsep}{8pt}
\resizebox{1.0\linewidth}{!}{
\begin{tabular}{lcccccc}
\toprule 
&\multicolumn{2}{c}{MIT-States}&\multicolumn{2}{c}{UT-Zappos}&\multicolumn{2}{c}{C-GQA}\\
\cmidrule(lr){2-3} \cmidrule(lr){4-5} \cmidrule(lr){6-7}
\textit{Acc}($\%$) & Attr  & Obj  & Attr  & Obj& Attr & Obj  \\
\midrule
Random& 1.13&0.47&5.74&8.14&0.25&0.05\\
Attr Branch& 26.3&3.87 &22.3&14.9&3.64&2.36 \\
Obj Branch&6.3 &45.2 &7.69&63.2&2.12&17.3\\
Comp Branch& 21.3&51.3 &21.6&58.9 &9.72&39.8\\
\hline
\end{tabular}
}
\label{tab:supp:leakage}
\end{table}

\subsection{Disentanglement Analysis}
We further refine our analysis by examining the role of the visual disentangler itself. In our initial cross-branch probe, we observe that on certain datasets the attribute and object branches can even become less predictive when the disentangler suppresses information too aggressively, suggesting that disentanglement may sometimes remove useful cues together with entangled ones. To better understand whether the disentangler remains beneficial overall, we conduct an additional controlled study: using the same backbone and training protocol, in Tab.~\ref{tab:supp:disentangler}, we compare (i) \textbf{w/ Disentangler}, the standard multi-path baseline~\cite{huang2024troika}, against (ii) \textbf{w/o Disentangler}, a variant in which we remove the disentangler.

Across all datasets, removing the disentangler leads to clear drops in HM and AUC, indicating that feature disentanglement, although imperfect and not strictly factor-pure, still provides meaningful structural benefits for compositional generalization.

In summary, the cross-branch “leakage’’ accuracies remain far above random under both settings, reinforcing our central observation that \emph{perfect} attribute–object separation is difficult to achieve in practice. This aligns with the motivation of our main paper: rather than pursuing an idealized, fully factorized representation, we embrace the remaining entanglement and turn it into a resource. Leakage-Guided Augmentation is therefore designed not to eliminate leakage, but to harness the surviving cross-factor signals as additional supervision, reflecting the philosophy that residual correlations, when properly guided, can be leveraged rather than suppressed.

\begin{table}[h]
\centering
% \vspace{-5pt}
\caption{Comparison between Troika and Troika without disentangler in MIT-States~\cite{mit}, UT-Zappos~\cite{ut} and C-GQA~\cite{cgqa} on HM and AUC under the closed-world setting.}
\setlength{\tabcolsep}{8pt}
\resizebox{1.0\linewidth}{!}{
\begin{tabular}{lcccccc}
\toprule 
&\multicolumn{2}{c}{MIT-States}&\multicolumn{2}{c}{UT-Zappos}&\multicolumn{2}{c}{C-GQA}\\
\cmidrule(lr){2-3} \cmidrule(lr){4-5} \cmidrule(lr){6-7}
 Disentangler& HM & AUC & HM & AUC & HM & AUC \\
\midrule
\textit{w}&39.2 &22.1 & 55.4&41.8&29.7&12.5 \\
\textit{w/o}& 36.8 & 20.4 & 51.4 & 37.5 & 28.0 & 11.2  \\
\hline
\end{tabular}
}
\label{tab:supp:disentangler}
\end{table}

\section{Additional Analysis}
\label{sec:supp:analysis}
\subsection{Additional Component Analysis}
We provide additional component analysis to quantify the contribution of each module: \emph{Primitive Flows (Flows)}, \emph{Composer}, and \emph{Leakage-Guided Augmentation (LG-Aug)} under the \textit{\textbf{open-world}} setting. 
As shown in Tab.~\ref{tab:supp:openWorldAblation}, we report results on MIT-States~\cite{mit}, UT-Zappos~\cite{ut}, and C-GQA~\cite{cgqa}, comparing against the Troika~\cite{huang2024troika} baseline.
Starting from the baseline, we incrementally add our components and make four observations.
\textbf{First}, introducing \emph{Primitive Flows} (Row (1)) delivers a consistent boost, especially on seen accuracy, indicating that explicit transport toward textual endpoints helps beyond token-level composition.
\textbf{Second}, adding \emph{Leakage-Guided Augmentation} on top of Flows (Row (2)) further improves robustness on unseen pairs, yielding higher HM and AUC by leveraging residual cross-branch cues.
\textbf{Third}, alternatively replacing LG-Aug with the \emph{Composer} (Flows+Composer, Row (3)) also stabilizes recognition by providing an explicit combination rule in the embedding space, improving HM and AUC over Flows alone.
\textbf{Fourth}, combining all components (\textbf{Ours}) achieves the best performance across all datasets and metrics, demonstrating that explicit velocity composition and leakage-guided supervision are complementary to flow learning in the open-world regime.

\begin{table*}[h]
\small
\centering
\caption{Ablations. The results regarding the different components in our \textit{FlowComposer} on MIT-States~\cite{mit}, UT-Zappos~\cite{ut} and C-GQA~\cite{cgqa} under the open-world setting.}
\setlength\tabcolsep{5.0pt}
\resizebox{1.0\linewidth}{!}{
\begin{tabular}{cccc|cccc|cccc|cccc}
\toprule \hline
&\textit{Flows} &\textit{Composer} & \textit{LG-Aug} & \multicolumn{4}{c|}{MIT-States~\cite{mit}} & \multicolumn{4}{c|}{UT-Zappos~\cite{ut}}& \multicolumn{4}{c}{C-GQA~\cite{cgqa}} \\
&(\S~4.1) & (\S~4.2) & (\S~4.3) & Seen & Unseen & HM & AUC & Seen & Unseen & HM & AUC & Seen & Unseen & HM & AUC\\
\midrule
Baseline&\xmark & \xmark & \xmark & 50.3&17.5&19.0&6.8& 65.8&61.0&46.6&32.7& 40.8  &8.6 &11.7 &2.9\\
(1)&\cmark & \xmark & \xmark & 50.1 & 18.0&19.6&7.0&68.5 & 60.8&49.3 & 33.8& 41.2&9.0&12.0&3.2 \\
(2)&\cmark & \xmark & \cmark & 49.7&18.9&20.2&7.3 &69.1&\textbf{61.5}&50.1&34.7&42.2&10.0&12.3&3.3  \\
(3)&\cmark & \cmark & \xmark &49.6&18.8&20.0&7.2 &69.8&61.1&50.7&34.9&41.9&\textbf{10.3}&12.2&3.2\\
\rowcolor{my_blue}Ours&\cmark & \cmark & \cmark & \textbf{50.4}&\textbf{19.0}&\textbf{20.3}&\textbf{7.5}& \textbf{70.1}&61.2&\textbf{51.0}&\textbf{35.5} &\textbf{43.5} & 10.2 & \textbf{12.6}&\textbf{3.5}\\
\bottomrule
\end{tabular}}
\label{tab:supp:openWorldAblation}
\vspace{-10pt}
\end{table*}

\subsection{Computational Costs}
Tab.~\ref{tab:comp_cost} reports the inference-time comparison between two baselines (CSP~\cite{nayaklearning} and Troika~\cite{huang2024troika}) and our \textit{FlowComposer}. We report the average latency (ms) per image over the entire test set.
Our method increases latency only slightly (\eg, from $17.6$ to $19.2$ ms per image on MIT-States~\cite{mit}, $11.0$ to $12.7$ ms per image on UT-Zappos~\cite{ut}, and $69.8$ to $74.4$ ms per image on C-GQA~\cite{cgqa} for Troika~\cite{huang2024troika}).
This modest cost stems from our one-step transport scheme and lightweight Composer, indicating that the performance gains come with negligible computation.

\begin{table}[h]
\centering
\caption{Inference time (ms/image) comparison between two baselines and ours.}
\setlength{\tabcolsep}{13pt}
\resizebox{1.0\linewidth}{!}{
\begin{tabular}{lccc}
\toprule
Method & MIT-States & UT-Zappos & C-GQA \\
\midrule
 CSP & 6.6&10.3&20.2\\
~+ Ours&9.3&12.5& 24.6 \\
 Troika & 17.6 & 11.0 & 69.8 \\
~+ Ours & 19.2 & 12.7 & 74.4 \\
\bottomrule
\end{tabular}
}
\label{tab:comp_cost}
\end{table}

\subsection{Composition Stepsize Analysis}
We introduce the hyperparameter stepsize $h$ of the one-step composition in Sec.~4.2.
As shown in the main paper Tab.~3, we evaluate $h \in \{0.1, 0.5, 1.0\}$ on MIT-States~\cite{mit} and UT-Zappos~\cite{ut} under the closed-world setting.
The optimal $h$ differs across datasets: UT-Zappos~\cite{ut} favors a larger step ($h{=}1.0$), whereas MIT-States~\cite{mit} prefers smaller steps.
We attribute this to dataset-specific composition geometry: UT-Zappos contains fewer, more separated compositions, where a longer step moves features closer to their text targets; in contrast, the denser composition spaces of MIT-States make large steps prone to overshooting and error accumulation.
Accordingly, we adopt dataset-specific choices of $h$ in all experiments.

Here, we provide an additional ablation study on C-GQA~\cite{cgqa}. As shown in Tab.~\ref{tab:supp:h}, the trend on C-GQA~\cite{cgqa} mirrors the behavior observed in MIT-States~\cite{mit} and UT-Zappos~\cite{ut}. A smaller composition stepsize ($h{=}0.1$) yields the best performance across all four metrics on both the validation and test sets. Larger stepsizes ($h{=}0.5$ and $h{=}1.0$) consistently reduce HM and AUC, indicating that overly long updates tend to overshoot the target direction in C-GQA’s dense composition space. The close alignment between validation and test results further confirms that the stepsize sensitivity is stable and not due to overfitting or evaluation noise. This provides additional support that different datasets exhibit distinct composition geometries, and that adopting dataset-specific stepsizes leads to more reliable one-step transport.

\begin{table}[ht]
\centering
\vspace{-5pt}
\caption{Results regarding stepsize $h$ (\S~4.2) on the validation and test set under the closed-world setting for C-GQA~\cite{cgqa}.}
\setlength{\tabcolsep}{3mm}
\resizebox{0.45\textwidth}{!}{
\begin{tabular}{lcccccccc}
\toprule 
& \multicolumn{4}{c}{Validation} & \multicolumn{4}{c}{Test}\\
\cmidrule(lr){2-5} \cmidrule(lr){6-9}
$h$ & S & U & H & A & S & U & H & A \\
\midrule
\rowcolor{my_blue}\textbf{0.1} & \textbf{44.6} & \textbf{40.6} & \textbf{32.9} & \textbf{15.5} & \textbf{44.8} & \textbf{40.7} & \textbf{34.0} & \textbf{15.9}\\
0.5 & 44.6 & 39.9 & 32.6 & 15.2 & 43.8 & 40.0 & 32.9 & 15.1 \\
1.0 & 42.0 & 36.7 & 29.8 & 13.0 & 40.3 & 36.9 & 29.7 & 12.7 \\
\bottomrule
\end{tabular}
}
\label{tab:supp:h}
\end{table}

% \begin{table}[ht]
% \centering
% \vspace{-5pt}
% \caption{Results regarding stepsize $h$ (\S~4.2) on the validation and test set under closed-world setting for C-GQA~\cite{cgqa}.}
% \setlength{\tabcolsep}{3mm}
% \resizebox{0.45\textwidth}{!}{
% \begin{tabular}{clcccccccc}
% \toprule 
% &&\multicolumn{4}{c}{Validation}&\multicolumn{4}{c}{Test}\\
% \cmidrule(lr){3-6} \cmidrule(lr){7-10}
%  & $h$ & S & U & H & A & S & U & H & A \\
% \midrule
%  \multirow{3}{*}{\rotatebox[origin=c]{90}{\textit{C-GQA}}}
%  \rowcolor{my_blue}& \textbf{0.1}  & \textbf{44.6} & \textbf{40.6} & \textbf{32.9} & \textbf{15.5 }& \textbf{44.8} & \textbf{40.7}& \textbf{34.0} & \textbf{15.9}\\
%  & 0.5 & 44.6 & 39.9 & 32.6 & 15.2 &  43.8& 40.0 & 32.9 & 15.1 \\
%  & 1.0 & 42.0 & 36.7&29.8 & 13.0 & 40.3 & 36.9 & 29.7 & 12.7  \\
% \bottomrule
% \end{tabular}
% }
% \label{tab:supp:h}
% \end{table}

\section{Stability Analysis}
\label{sec:supp:5runs}
Following standard practice~\cite{nayaklearning, huang2024troika}, we further evaluate the stability of our method over $5$ random seeds. 
Tab.~\ref{tab:sipp:5runs} reports the mean and standard deviation across five independent runs on all three benchmarks. 
The variances remain consistently small, and the performance gains over both baselines persist under this multi-run setting. 
Notably, even on the challenging C-GQA dataset, \textit{FlowComposer} delivers stable and reproducible improvements. 
These results demonstrate that \textit{FlowComposer} is not only effective but also robust to different initializations.

\begin{table*}[h]
% \vspace{-0.5em}
  \caption{Performance on MIT-States~\cite{mit}, UT-Zappos~\cite{ut}, and C-GQA~\cite{cgqa}, averaged over 5 random seeds with corresponding standard deviations. * marks results taken from the respective original publications.
}
  \label{tab:sipp:5runs}
  \centering
  \setlength{\tabcolsep}{2.0pt}{
  \resizebox{1\linewidth}{!}{
\begin{tabular}{l|cccc|cccc|cccc}
\toprule
\hline
&\multicolumn{4}{c|}{MIT-States~\cite{mit}}& \multicolumn{4}{c|}{UT-Zappos~\cite{ut}}&\multicolumn{4}{c}{C-GQA~\cite{cgqa}}\\
% \cmidrule(lr){3-6} \cmidrule(lr){7-10} \cmidrule(lr){11-14} 
Method&Seen&Unseen&HM&AUC&Seen&Unseen&HM&AUC&Seen&Unseen&HM&AUC\\
\hline
CSP$^*$~\cite{nayaklearning}\scriptsize \textcolor{gray}{[ICLR23]} & 46.6\tiny$\pm0.1$& 49.9\tiny$\pm0.1$ & 36.3\tiny$\pm0.1$&19.4\tiny$\pm0.1$ & 64.2\tiny$\pm0.7$&66.2\tiny$\pm1.2$&46.6\tiny$\pm1.2$&33.0\tiny$\pm1.3$&28.8\tiny$\pm0.1$&26.8\tiny$\pm0.1$&20.5\tiny$\pm0.1$&6.2\tiny$\pm0.0$\\
\cellcolor{my_blue}\textbf{+FlowComposer}&\cellcolor{my_blue}48.4\tiny$\pm0.2$&\cellcolor{my_blue}50.4\tiny$\pm0.1$&\cellcolor{my_blue}37.8\tiny$\pm0.2$&\cellcolor{my_blue}20.7\tiny$\pm0.1$&\cellcolor{my_blue}66.9\tiny$\pm0.4$&\cellcolor{my_blue}68.0\tiny$\pm0.7$&\cellcolor{my_blue}51.4\tiny$\pm0.7$&\cellcolor{my_blue}38.2\tiny$\pm0.8$&\cellcolor{my_blue}29.0\tiny$\pm0.1$&\cellcolor{my_blue}31.0\tiny$\pm0.2$&\cellcolor{my_blue}23.1\tiny$\pm0.2$&\cellcolor{my_blue}7.8\tiny$\pm0.2$ \\
Troika$^*$~\cite{huang2024troika}\scriptsize \textcolor{gray}{[CVPR24]}& 49.0\tiny$\pm0.4$&53.0\tiny$\pm0.2$&39.3\tiny$\pm0.2$&22.1\tiny$\pm0.1$& 66.8\tiny$\pm1.1$&73.8\tiny$\pm0.6$&54.6\tiny$\pm0.5$&41.7\tiny$\pm0.7$&41.0\tiny$\pm0.2$&35.7\tiny$\pm0.3$&29.7\tiny$\pm0.2$&12.4\tiny$\pm0.1$ \\
\cellcolor{my_blue}\textbf{+FlowComposer}& \cellcolor{my_blue}51.7\tiny$\pm0.3$&\cellcolor{my_blue}53.1\tiny$\pm0.1$&\cellcolor{my_blue}40.2\tiny$\pm0.1$&\cellcolor{my_blue}23.4\tiny$\pm0.1$& \cellcolor{my_blue}71.5\tiny$\pm0.6$&\cellcolor{my_blue}75.0\tiny$\pm0.3$&\cellcolor{my_blue}58.8\tiny$\pm0.4$&\cellcolor{my_blue}46.7\tiny$\pm0.3$&\cellcolor{my_blue}44.8\tiny$\pm0.1$&\cellcolor{my_blue}40.6\tiny$\pm0.1$&\cellcolor{my_blue}34.1\tiny$\pm0.1$&\cellcolor{my_blue}15.9\tiny$\pm0.1$ \\
\bottomrule
\end{tabular}
}
}
\vspace{-10pt}
\end{table*}

\section{Semantic Analysis of Composer}
\label{sec:supp:composer}
\begin{figure*}[h]
    \centering
    \includegraphics[width=1\textwidth]{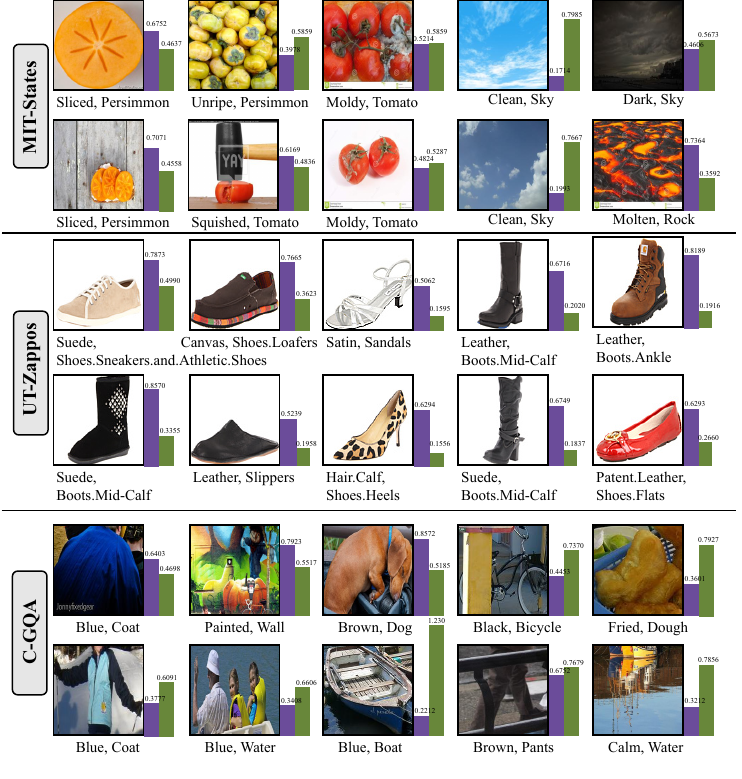}
    \caption{The purple bar \colorbox[HTML]{6B4C9A}{\phantom{\rule{1em}{1em}}} denotes the value of $\hat{a}$ and the green bar \colorbox[HTML]{68873A}{\phantom{\rule{1em}{1em}}} denotes the value of $\hat{b}$, which are the coefficients for attribute and object velocities respectively.
    }
    \label{fig:supp:ab}
\end{figure*}

To better understand the behavior of our \textit{Composer} and examine whether the predicted combination coefficients carry semantic meaning, we visualize the predicted attribute and object weights $(\hat{a},\hat{b})$ for correctly classified samples across the three datasets using the \textit{FlowComposer} based on Troika~\cite{huang2024troika}. Recall that $\hat{a}$ and $\hat{b}$ correspond to the contributions of the attribute and object velocities when composing the final composition velocity. Importantly, our intention is not to enforce that the ground-truth composition velocity must lie exactly in the span of the primitive velocities, $\mathrm{span}\{v_{\text{attr}}, v_{\text{obj}}\}$. Instead, the \textit{Composer} aims to predict an approximate combination that produces a composition velocity pointing toward the desired semantic direction. As shown in Fig.~\ref{fig:supp:ab}, the learned coefficients indeed correlate with the visual prominence of attribute and object cues, demonstrating that the \textit{Composer} adapts its weighting based on the evidence present in the image.

Across the visualizations, we observe clear and consistent semantic tendencies. When an attribute is visually salient, such as distinctive material patterns, strong color cues, or highly recognizable texture changes, the predicted attribute coefficient $\hat{a}$ increases. In contrast, when the attribute is abstract, weakly expressed, or visually ambiguous, the \textit{Composer} assigns a smaller value to $\hat{a}$, relying more heavily on the object velocity. Similarly, when the object identity is visually clear and structurally well preserved, the object coefficient $\hat{b}$ becomes dominant; however, if the object shape is partially occluded, deformed, or visually degraded, the Composer reduces $\hat{b}$ and compensates through attribute evidence. These trends appear consistently across MIT-States~\cite{mit}, UT-Zappos~\cite{ut}, and C-GQA~\cite{cgqa}, despite the substantial differences in attribute style, object granularity, and visual complexity among the datasets.
For example, in MIT-States~\cite{mit}, attributes such as \texttt{sliced} or \texttt{squished} exhibit strong, localized visual patterns, leading to larger $\hat{a}$, whereas abstract attributes like \texttt{unripe}, \texttt{clean} or \texttt{dark} produce much smaller coefficients due to their weak visual expression. Similarly, in C-GQA~\cite{cgqa}, samples with vivid attribute cues (\eg, \texttt{blue coat}, \texttt{brown dog} or \texttt{painted wall}) yield higher $\hat{a}$, while cases where the attribute is subtle or occluded (\eg, \texttt{blue water} \texttt{black bicycle}, or \texttt{Fried Dough}) shift the weighting toward the object, resulting in larger $\hat{b}$.
For the UT-Zappos~\cite{ut} dataset, the attribute typically describes material or surface appearance (\eg, \texttt{leather}, \texttt{satin}, \texttt{canvas}), which is prominently expressed on the shoe surface, while the object denotes the shoe type, whose structural cues are often less visually distinctive. As a result, the attribute coefficient $\hat{a}$ consistently dominates across UT-Zappos samples, reflecting the stronger visual salience of material-based attributes.

The sample-level behaviors in Fig.~\ref{fig:supp:ab} further highlight the interpretability of the predicted coefficients. When attribute cues are strong or highly distinctive, the Composer increases the corresponding attribute weight, whereas objects with vivid, easily recognizable visual characteristics naturally lead to larger object coefficients. 
For example, in the two \texttt{blue coat} samples from C-GQA~\cite{cgqa}, the first image exhibits a much more prominent blue coloration, leading to a larger attribute coefficient, whereas in the second image the coat shape is clearly visible but the blue cue is relatively weak, resulting in a higher object coefficient $\hat{b}$.
Similarly, for the two \texttt{moldy tomato} samples from MIT-States~\cite{mit}, the upper image contains more pronounced moldy regions and multiple clearly visible tomatoes, which yields larger values for both the attribute and the object coefficients compared to the lower sample, where the mold is less salient and fewer tomatoes are present.

These observations collectively reinforce the core motivation behind our \textit{Composer}: different composites and even different samples of the same composite exhibit varying visual contributions from the attribute and the object, so a fixed weighting scheme (as used in prior multi-branch methods such as Troika~\cite{huang2024troika}) cannot adequately capture such diversity. In contrast, our learned coefficients provide sample-specific balancing of primitive velocities, allowing the model to adapt its reliance on attribute or object information according to their respective visual strengths. This semantic analysis therefore supports our design choice of learning explicit, data-dependent combination coefficients and demonstrates that the \textit{Composer} produces weights that align with human-interpretable visual cues rather than applying uniform or heuristic mixing rules.

\section{Social Impacts}
\label{sec:supp:impact}
This paper presents a feasible approach to compositional zero-shot learning, enabling models to recognize novel attribute–object compositions from known primitives, which may potentially benefit a variety of applications such as retrieval, assistive perception, or visual reasoning in open-world environments. 
However, there remains a potential risk associated with the reliance on pretrained vision–language models and benchmark datasets. 
In particular, our method may inherit and amplify dataset biases—such as skewed attribute distributions or culturally sensitive descriptors—which can propagate through the compositional mechanism. 
Such amplified biases may in turn lead to incorrect or inappropriate composition predictions in high-stakes scenarios, where misidentifying attributes or objects could result in harmful downstream decisions. 
Additionally, to mitigate potential negative social impacts, the development of rigorous bias-auditing tools, transparent evaluation practices, and responsible deployment protocols is crucial to ensure that such models do not propagate harmful biases or influence high-stakes decisions without proper safeguards.


\end{document}